\documentclass[journal,10pt]{IEEEtran}
\ifCLASSOPTIONcompsoc

\usepackage[nocompress]{cite}
\else

\usepackage{cite}
\fi

\ifCLASSINFOpdf
\usepackage[pdftex]{graphicx}
\else
\usepackage[dvips]{graphicx}
\fi

\usepackage{cite}

\usepackage{graphicx}
\usepackage{textcomp}

\usepackage{xcolor}

\usepackage{amsfonts}
\usepackage{amssymb}
\usepackage[cmex10]{amsmath}
\usepackage{amsthm}
\usepackage{cases}
\usepackage{subfigure}
\usepackage{booktabs}
\usepackage{colortbl}
\usepackage{tabularx}
\usepackage{algorithm}
\usepackage{algorithmic}
\usepackage{color}
\usepackage{subfigure}
\usepackage{url}
\usepackage{setspace}
\usepackage{caption}
\usepackage{ragged2e}
\begin{document}

\title{Towards Secure and Efficient Data Scheduling for Vehicular Social Networks}
	
\author{Youhua Xia,~\IEEEmembership{Member,~IEEE},
	Tiehua Zhang,~\IEEEmembership{Member,~IEEE},
	Jiong Jin,~\IEEEmembership{Senior Member,~IEEE},
	Ying He,~\IEEEmembership{Member,~IEEE}, and
	Fei Yu,~\IEEEmembership{Fellow,~IEEE}
	
\thanks{Corresponding author: Tiehua Zhang.}
\thanks{Youhua Xia and Fei Yu are with the Guangdong Laboratory of Artificial Intelligence and Digital Economy (SZ), Shenzhen, P.R. China, 518107, Fei Yu is also with the College of Computer Science and Software Engineering, Shenzhen University, Shenzhen, P.R. China, 518060. E-mail: xiayouhua@gml.ac.cn, yufei@gml.ac.cn.}
\thanks{Tiehua Zhang is with the College of Electronics and Information Engineering, Tongji University, Shanghai, P.R. China. E-mail: tiehuaz@tongji.edu.cn}
\thanks{Jiong Jin is with the School of Science, Computing and Engineering Technologies, Swinburne University of Technology, Melbourne VIC 3122, Australia. E-mail: jiongjin@swin.edu.au}
\thanks{Ying He is with the College of Computer Science and Software Engineering, Shenzhen University, Shenzhen, P.R. China, 518060. E-mail: heying@szu.edu.cn. 	}

%\thanks{F. Richard Yu is with the Guangdong Laboratory of Artificial Intelligence and Digital Economy (SZ), Shenzhen, P.R.China, 518107. E-mail: Richard.Yu@Carleton.ca.}

\thanks{Manuscript received XXX, XX, 2023; revised XXX, XX, 2024.}}

\markboth{IEEE Transactions on Vehicular Technology,~Vol.~XX, No.~XX, XXX~2024}
{}

\maketitle

\IEEEcompsoctitleabstractindextext{

\begin{abstract}
 Efficient data transmission scheduling within vehicular environments poses a significant challenge due to the high mobility of such networks. Contemporary research predominantly centers on crafting cooperative scheduling algorithms tailored for vehicular networks. Notwithstanding, the intricacies of orchestrating scheduling in vehicular social networks both effectively and efficiently remain a formidable task. This paper introduces an innovative learning-based algorithm for scheduling data transmission that prioritizes efficiency and security within vehicular social networks. The algorithm first uses a specifically constructed neural network to enhance data processing capabilities. After this, it incorporates a Q-learning paradigm during the data transmission phase to optimize the information exchange, the privacy of which is safeguarded by differential privacy through the communication process. Comparative experiments demonstrate the superior performance of the proposed Q-learning enhanced scheduling algorithm relative to existing state-of-the-art scheduling algorithms in the context of vehicular social networks.
\end{abstract}

}

\maketitle
\IEEEdisplaynotcompsoctitleabstractindextext
\IEEEpeerreviewmaketitle
	
	\section{Introduction}

The development of modern cities has witnessed the significant growth of vehicles and the increasing time spent driving vehicles for daily transportation. Therefore, communication between vehicles has attracted widespread attention from researchers and industrial practitioners, and the vehicular social network has also become an essential part of modern networks. Collaborative vehicular social networks are increasingly crucial to the development of the V2X (Vehicle to Everything) communication paradigm~\cite{xia2020cluster, ahmed2018cooperative}. To make communication between vehicles more efficient and secure, it is necessary to enable efficient collaboration between vehicles while maintaining the privacy of data exchange. Cooperative data scheduling finds the best solution for efficient and secure data communication between vehicles~\cite{luo2020collaborative, li2020deep}.

%\red{The communication in vehicular social networks can be summarized as V2X. Due to the heterogeneity and dynamics of the vehicular social network, it is still difficult to achieve ultra-low latency and large-scale communication between vehicles~\cite{tang2019future}. To have a better communication experience, data needs to be efficiently and securely transmitted among vehicles. For example, a reinforcement learning-based routing protocol is designed to achieve better data packet forwarding in a multi-access vehicular environment~\cite{wu2020collaborative}. Considering the vehicle's real-time service requirements, it is necessary to adaptively dispatch the vehicle's dynamic information. To provide real-time services, Sharma et al. \cite{sharma2020adps} proposed a priority service-based data scheduling algorithm, which uses fuzzy logic to estimate the deadline. However, it is difficult to meet the high efficiency and security requirements of data transmission between vehicles at the same time.}

Due to the highly dynamic characteristics of vehicles, large-scale ultra-low-latency communication in vehicular social networks becomes very difficult~\cite{tang2019future}. 
To provide real-time data services for vehicles, Wu \emph{et al.}~\cite{wu2020collaborative} use reinforcement learning to implement data packet forwarding of routing protocols in multi-access vehicle environments, and Sharma\cite{sharma2020adps} and Awasthi propose a data scheduling algorithm based on priority services, estimating the deadline through fuzzy logic and adaptively scheduling dynamic vehicle information. To improve the reliability of data transmission, Xia \emph{et al.} \cite{9794641} proposed an incentive-driven privacy-preserving data scheduling for the Internet of Vehicles, using an incentive mechanism to achieve efficient data transmission scheduling. To improve the network's security, Zhang \emph{et al.} \cite{9347691} designed a weight-based integrated machine learning algorithm and established an intrusion detection model based on multi-objective optimization by identifying abnormal messages. However, the above-mentioned works fail to simultaneously meet the requirements of high efficiency and security of data transmission when dealing with large amounts of vehicles and data.
%To improve the communication experience, reinforcement learning is used to implement packet forwarding of routing protocols in a multi-access vehicle environment~\cite{wu2020collaborative}, making data transmission secure and efficient. To provide real-time services to vehicles, Sharma and Awasthi \cite{sharma2020adps} proposed a data scheduling algorithm based on priority services, which uses fuzzy logic to estimate deadlines and adaptively schedules dynamic information of vehicles. However, due to the increasing number of vehicles and the amount of data, it is still difficult to simultaneously meet the requirements of high efficiency and security of data transmission between vehicles.

In vehicular social networks, providing real-time data services for vehicles is a challenging problem. First of all, the highly dynamic nature of the vehicle makes the communication time between the vehicles very short. Then, data processing becomes increasingly difficult with limited resources, and data processing on vehicles is expected to be more efficient in this highly dynamic environment. Finally, the data security issue among vehicles raises another concern. These challenges have brought great difficulties when implementing real-time, high-quality service to vehicles. 

%To this end, a novel approach is proposed to enhance the efficiency and intelligence of data transmission services in vehicular networks. The key focus is on safeguarding data privacy during transmission, achieved through the integration of differential privacy into the data communication process. The proposed algorithm incorporates Q-learning for optimizing data transmission, a multi-layer perceptron for efficient data processing, and differential privacy techniques for ensuring privacy preservation. Firstly, Q-learning is employed to enhance the efficiency of data transmission in vehicle-to-vehicle communication. Subsequently, the multi-layer perceptron is utilized to narrow down the scope of data processing, thereby accelerating data acquisition speed. Finally, differential privacy measures are employed to protect the privacy of transmitted data, ensuring both efficiency and security of the service. Collectively, these contributions address the challenges of intelligent and secure data transmission in vehicular networks.

To this end, we propose a data scheduling algorithm based on Q-learning to achieve efficient data communication and processing while ensuring security. The proposed DSQL algorithm first considers factors such as the amount of data processing and the probability of encountering obstacles during transmission to improve the design of the reward function and combines energy consumption optimization to make communication more efficient Then, it performs sine function mapping on the input data to reduce the data range and improve the data processing efficiency of the multi-layer perceptron. Finally, it sets effective interference conditions by combining pseudonym entropy and uses differential privacy mechanisms to protect the reward values and action information of vehicles.

% this paper proposes a learning-based data transmission scheduling to enable intelligent and efficient data transmission services in a vehicular network. To protect data privacy during transmission, differential privacy is merged into the data communication process to ensure privacy-preserving purposes. Specifically, the algorithm contains data communication based on Q-learning, data processing based on multi-layer perceptron, and data privacy protection. Firstly, Q-learning is used to improve the efficiency of data transmission during communication between vehicles; then, the scope of data processing is narrowed through the multi-layer perceptron, so that the speed of data acquisition is faster; finally, the privacy protection of data is carried out through differential privacy, so that the service is both efficient and secure.

The main contributions of this work are summarized as:
\begin{itemize}
	\item We propose a data scheduling algorithm based on Q-learning, which uses the amount of locally processed data to measure the size of the reward value, and optimizes energy consumption based on the distance between nodes, thereby improving the efficiency of data communication between dynamic nodes within the vehicular social network.
 %We propose a Q-learning based data scheduling algorithm, which integrates energy consumption optimization to enhance the efficiency of data communication within vehicular social networks.
 
	\item We use the sine function to map the data to a smaller range for processing through the design of the activation function of the multi-layer perceptron, thereby improving the data processing capabilities. Then, under privacy constraints, differential privacy is used to protect the vehicle's reward value and action information.
 %We use a smaller data range to enable the multi-layer perceptron to process data efficiently, then obtain the processed results, and use differential privacy to protect the privacy of the data.
 
	\item We verify the performance of the algorithm proposed in this paper through simulation experiments and conduct a comparative analysis with existing algorithms. Experimental results show that in vehicular social networks, the algorithm proposed in this paper is better than existing data scheduling algorithms in terms of accuracy, travel expenses, connectivity degree, transmission delay, and probability of privacy leakage.
 %We conduct simulated experiments to verify the performance of the proposed method while comparing it to the existing algorithms. The experimental results show that the algorithm proposed in this paper performs better than other existing scheduling algorithms in vehicular social networks.

\end{itemize}

The remainder of the paper is structured as follows. Section \ref{sec:related} discusses related work. Section \ref{section:system} presents the system model and problem formulation. Section \ref{section:algorithm} proposes the Q-learning-based data scheduling algorithm. Experimental results are conducted in Section \ref{section:results analysis}. Section \ref{section:Discussion} discusses machine learning-based data scheduling for vehicular networks. Section \ref{section:conclusion} concludes this paper.

	\section{Related Work}\label{sec:related}
%\red{privacy-driven scheduling related work}
% In this section, data service scheduling, data scheduling based on edge computing and fog computing, and data scheduling based on distributed computing are first studied. The data communication services, reliable scheduling based on data security, and collaborative scheduling algorithms are studied to make data transmission more efficient. 
%To enable vehicles to transmit data securely and quickly, data scheduling based on collaborative learning is researched in vehicular social networks. Due to the high-speed mobility of vehicles, 
% to make data transmission more efficient
%to settle the security of data transmission
We briefly discuss the related work from the following aspects: data service scheduling, data scheduling based on edge computing and fog computing, distributed computing-based data scheduling, data communication services, data security-based reliable scheduling, cooperative scheduling algorithm, privacy-driven scheduling, and secure data scheduling.

\subsection{Data Service Scheduling}

Based on fuzzy logic and priority, Sharma and Awasthi\cite{sharma2020adps} proposed a data scheduling algorithm to provide real-time information services. Priority requests are stored in a multi-level queue according to the degree of urgency for vehicles. Ning \emph{et al.}\cite{ning2019deep} have built an intelligent offloading framework for the vehicular social network in the 5G environment by using spectrum resources. The cost minimization problem is explained through delay constraints, and the problem is decomposed into two sub-problems. Zhang \emph{et al.}\cite{zhang2017sovcan} developed a safety-oriented vehicle controller area network based on a software-defined network method. To achieve the goal of ensuring traffic safety, the driver's emotions are identified by monitoring the driver's physical and psychological state. Dai \emph{et al.}\cite{dai2020multi} jointly utilize cloud computing and edge computing resources to balance the computing offload and workload in the vehicular social network. The distributed task allocation problem is solved by considering heterogeneous computing resources and uneven distribution of workloads, an algorithm based on multi-arm bandit learning is proposed, which is called utility table-based learning.

\subsection{Data Scheduling Based on Edge Computing and Fog Computing}
Wang \emph{et al.}\cite{wang2018enabling} focused on the collaboration between different edge computing nodes and proposed a collaborative vehicular edge computing framework. Support more scalable vehicle services and applications through vertical and horizontal collaboration, and discuss the technologies that support the framework. Dong \emph{et al.}\cite{dong2017enhancing} researched the practical 5G-enabled intelligent collaborative vehicular social network architecture, considering various technical characteristics of 5G networks and different mobile scenarios of vehicles, and using this architecture to assure bandwidth and communication reliability. To deploy smart data computing strategies, Lin \emph{et al.}\cite{lin2020distributed} introduced fog computing into the vehicular social network and proposed an architecture that supports software-defined networks, which includes a network layer, a fog layer, and a control layer. A hybrid scheduling algorithm aims to solve the multiple time-constrained vehicle application scheduling problems. For efficient transmission scheduling, Zhang \emph{et al.}\cite{zhang2018artificial} used the deep Q-learning method to design the best transmission scheduling scheme in cognitive vehicular social networks, and communication resources are leveraged to minimize transmission costs. Considering the characteristics of spectrum resources, an efficient learning algorithm is proposed for optimal scheduling strategy.

\subsection{Distributed Computing Based Data Scheduling}
To complete real-time services within a certain range, Dai \emph{et al.}\cite{dai2019learning} studied a service scenario based on mobile edge computing. Considering the characteristics of delay requirements, a distributed real-time service scheduling problem is proposed to maximize the service ratio. Subsequently, they studied a solution for computing offloading in a mobile edge computing-assisted architecture and studied task upload coordination, task migration, and heterogeneous computing among multiple vehicles. A probabilistic computing offloading algorithm is proposed to solve the collaborative computing based on queuing theory and minimize the delay in completing tasks\cite{9180064}. To reduce the occurrence of traffic accidents and raise traffic safety at the same time, Toutouch and Alba\cite{toutouh2018swarm} proposed a distributed congestion control strategy based on swarm intelligence to ensure the quality of network service while maintaining the level of channel usage. Xiong \emph{et al.}\cite{xiong2020reinforcement} proposed the architecture of intrusion detection and defense system for the vehicular social network, applying reinforcement learning to respond to the dynamic changes of vehicles, and making decisions based on the current state to obtain higher detection accuracy. Due to the sensitivity to security applications, the architecture is deployed in edge computing.

\subsection{Data Communication Services}
To summarize efficient and reliable vehicular communication from the perspective of the network layer, Peng \emph{et al.}\cite{peng2018vehicular}  first introduced the classification of vehicular communication, and then available communication technologies, network structures, and routing protocols are discussed. Finally, the challenges faced by the vehicular social network are identified. To grow the detection accuracy of the surrounding environment, Aoki \emph{et al.}\cite{aoki2020cooperative} proposed a collaborative perception scheme based on deep reinforcement learning. Through deep reinforcement learning, the data to be transmitted is selected and the network load is reduced. A collaborative intelligent vehicle simulation platform is developed, which integrates three software components: a traffic simulator, a vehicle simulator, and an object classifier. To efficiently manage the spectrum resources, Paul \emph{et al.}\cite{paul2015cooperative} proposed a centralized and distributed cooperative spectrum sensing system model for the vehicular social network. The system model is designed to minimize high-speed mobility and scarcity of spectrum by analyzing the decision-making fusion technology and using the updated theory. To elevate spectrum efficiency, Su \emph{et al.}\cite{su2020green} proposed a cloud vehicular wireless access network architecture, which includes centralized processing, cooperative radio, real-time cloud computing, and data compression functions. Based on this architecture, an efficient data compression method is proposed to meet the data compression requirements of the cellular vehicular social network.

\subsection{Data Security-Based Reliable Scheduling}
Posner \emph{et al.}\cite{posner2021federated} studied a new type of vehicular social network, the federal vehicular social network, which is a robust distributed vehicular network with better scalability and stability. Promote transactions and reduce malicious behaviors through the auxiliary blockchain system, and share data and models through the federal vehicular cloud. To reduce the delay and raise the service reliability of vehicle users, Li \emph{et al.}\cite{li2020collaborative} proposed an efficient method of computing offloading and server collaboration. Multiple edge servers are used to share the computing tasks of vehicles, reduce delay in parallel computing, and use collaboration between edges to reduce data transmission failures. Huang \emph{et al.}\cite{huang2021dynamic} analyzed in detail the architecture and communication scheduling algorithm of the vehicular network and proposed a dynamic priority strategy to balance the efficiency of data processing. This strategy completes the process of data sending, data transmission, and data receiving, and considers the feedback process to update the data priority. Zhang \emph{et al.}\cite{zhang2020v2x} built a software-defined network-assisted mobile edge computing network architecture for the vehicular network and solved the problem of V2X offloading and resource allocation through the best offloading decision-making and computing resource allocation scheme. The SDN controller is used to perceive the network status from a global perspective so that the vehicular network has higher efficiency and flexibility.

\subsection{Cooperative Scheduling Algorithm}
To make virtual networks cooperate, Wang \emph{et al.}\cite{WANG2019105557} used software to define network controllers, proposed a deep reinforcement learning method, and introduced a Markov model to take advantage of the heterogeneous performance of virtual networks. Based on this, a collaborative solution based on the asymmetric Nash protocol is proposed. Zeng \emph{et al.}\cite{zeng2017channel} proposed a scheduling strategy based on channel prediction to promote system throughput. Through the recursive least squares algorithm, the communication overhead of data distribution between vehicles is lower, and the scheduling efficiency is higher.

\subsection{Privacy-Driven Scheduling}
To prevent cyber criminals from exploiting the loopholes in information exchange between vehicles, Rathore \emph{et al.}\cite{rathore2022novel} proposed a trust-driven privacy method for efficient and secure transmission using encryption and steganography, thereby enhancing the data security of real-time communication between vehicles. To prevent life and property from being damaged, it is necessary to meet the requirements of performance parameters. To meet the trust index between vehicles and devices, Sharma and Kumar\cite{sharma2021third} proposed a security information management scheme for parameter evaluation, including average energy consumption, average hop count, attack intensity, error probability, etc.. Limbasiya \emph{et al.}\cite{9375496} used a one-way hash function and elliptic curve cryptography to propose a new secure and energy-saving message communication system MComIoV, and evaluated MComIoV and verified its robustness through security proof and analysis against various attacks. To solve the privacy leakage problem in mobile edge computing offloading, Pang \emph{et al.}\cite{pang2022towards} propose a novel online privacy-preserving computing offloading mechanism, OffloadingGuard, which minimizes the total cost of task computing while protecting privacy. To achieve the trade-off between user privacy and computational cost, a reinforcement learning-based offloading model is designed to adaptively determine a satisfactory perturbation offloading rate. To meet the different privacy requirements of edge applications, Zhu \emph{et al.}\cite{zhu2021learning} proposed a learning-authorized privacy protection scheme to adaptively perturb application data in multi-modal differential privacy. Then, take a case study to implement the scenario in edge cache service management.

\subsection{Secure Data Scheduling}
Geng \emph{et al.} \cite{10015040} proposed an overall framework based on reinforcement learning for Internet of Vehicles routing. Based on the Markov decision process, the routing problem was modeled as the optimization of constraint satisfaction problems. Under the constraints of satisfying peak and average delays, the optimal strategy for the constraint satisfaction problem is proved by extending the Q-learning algorithm, and the scalability of the framework is improved through decentralized implementation. To improve the service quality of generated artificial intelligence, Zhang \emph{et al.} \cite{10506539} proposed a multi-modal semantic perception framework, using multi-modal and semantic communication technology to build multi-modal content and improve the usability and efficiency of on-board systems. A resource allocation method based on deep reinforcement learning is proposed to improve the reliability of generative artificial intelligence V2V communication, and the research progress in the field of generative artificial intelligence car networking is discussed. Yao \emph{et al.} \cite{10016623}first developed a distributed Kalman filter algorithm to share state estimates among adjacent nodes to track attackers. Then, the communication channel and transmission power are designed while ensuring the communication quality, and an architecture based on the hierarchical deep Q network is developed to design the anti-eavesdropping power and channel selection. To simultaneously meet the throughput and transmission time requirements of the Internet of Vehicles, Li \emph{et al.} \cite{10012329}proposed a joint optimization problem of relay selection and transmission scheduling and solved the problem through concurrent scheduling of random relay selection and dynamic scheduling of joint relay selection. Ju \emph{et al.} \cite{10041957}proposed a joint secure offloading and resource allocation scheme based on deep reinforcement learning to improve the security performance and resource efficiency of multi-users in the Internet of Vehicles. Through the joint optimization of transmit power, spectrum selection, and computing resource allocation, the optimal decision-making of multi-agent collaboration is utilized, and the dual-deep Q learning algorithm is used to solve this optimization problem. Twardokus and Rahbari \cite{10058928} develop covert denial-of-service attacks that exploit vulnerabilities in the C-V2X security protocol. Low-duty cycle attacks will reduce the availability of C-V2X and increase the risk of vehicle driving. Then, develop attack detection and mitigation technology, which may use the new C-V2X functions of 3GPP Rel-17.

The above works try to ensure the efficiency and security of data transmission from the perspective of data service scheduling, distributed computing-based data scheduling, data communication services, data security-based reliable scheduling, cooperative scheduling algorithm, privacy-driven scheduling, and secure data scheduling, respectively. However, data is difficult to be processed faster, and it is challenging to acquire data efficiently. In this paper, we integrate the multi-layer perceptron with Q-learning and propose Q-learning-based data scheduling for the vehicular network.
	
	\section{System Model and Problem Formulation} \label{section:system}

In this section, we present the system model and formulate the data scheduling problem in vehicular networks.

\begin{figure}[!tbp]
	\centering
	\includegraphics[width=0.45\textwidth]{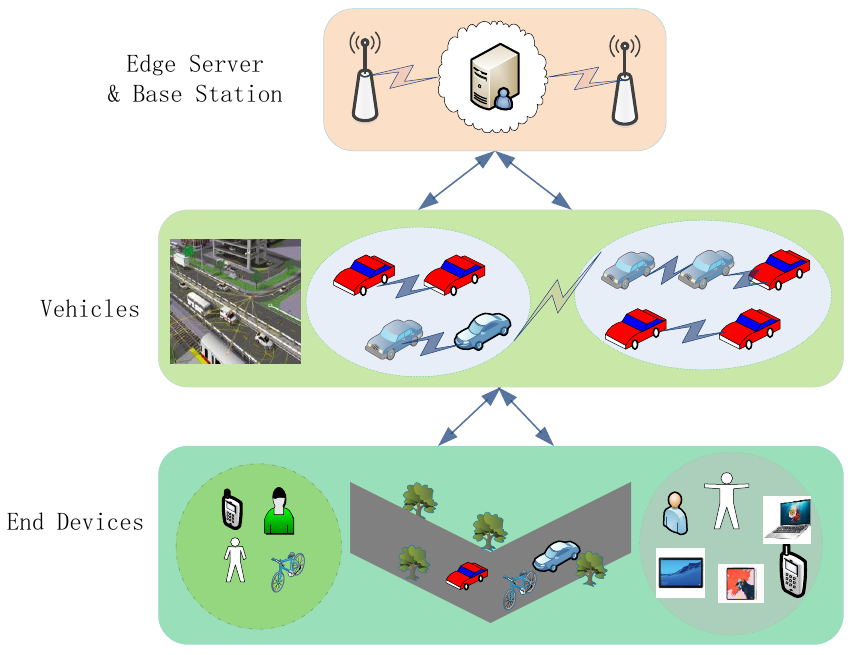}
	\centering{\caption{System model.}\label{figures:nrs}}	
\end{figure}

\begin{table}
	\caption{Summary of Notations}\label{SN}	\centering
	\par
	\footnotesize
	{
		\begin{tabular}{p{1.5cm}|p{6cm}}
			\hline
			\hline
			Notations & Descriptions  \\
			\hline
			$dn$ & The destination node    \\
			\hline
			$te$ & The communication type    \\
			\hline
			$LQ(c, m)$ & The value of link quality between node $c$ and $m$    \\
			\hline
			$\hat{Red}$ & The reward from the BS through the cellular interface   \\
		    \hline
		%	$\lambda$ & Average number of vehicles reaching dense region    \\
		%	\hline
			$PD_{th}$ & The delay requirement   \\
			\hline
			$T_{persistent}$ & The duration of time the vehicle stays in motion  \\
			\hline
			$CB_{ul}$ & The uplink bandwidth   \\
			\hline
			$CB_{dl}$ & The downlink bandwidth   \\
			\hline
			$PD_{bs}$ & The processing delay at the base station   \\
			\hline
			$B$ & The bandwidth of the channel    \\
			\hline
			$E_{n}^{\text {local }}$ & The energy consumption for local processing of vehicles $n$ during a certain period of time   \\
			\hline
			$D_{n}^{\text {local }}$ & The amount of data processed locally by the vehicle $n$ in a certain period of time   \\
			\hline
			$d^{-\vartheta}$ & The path loss   \\
			\hline
			%$l_{ti}$ & The distance deviation  \\
			%\hline
			$d$ & The distance between two nodes     \\
			\hline
			$\vartheta$ & The path-loss index    \\
			\hline
			$r_{n, k}^{t, V2I }$ & The data transmission rate from vehicle $n$ to RSU (Road Side Unit) $k$ in V2I   \\
			\hline
			$p_{n}^{\mathrm{tr}}$ & The transmission power of vehicle $n$    \\
			\hline
			$\omega_{0}$ & The white Gaussian noise power    \\
			\hline
			$d_{n, k}^{t}$ & In a certain period of time $t$, the distance between the vehicle $n$ and RSU $k$    \\
			\hline
			$E_{n}^{\mathrm{tr}}$ & The energy consumption of the vehicle $n$ to transmit data in a certain period of time    \\
			%	$\Omega\left(N_{i}, G\right)$ & The node $N_{i}'s$ neighbor set in graph $G$    \\
		%	\hline
		%	$NB_{m}$  &   The one-hop neighbor set of node $m$  \\
		%	\hline
		%	$\alpha$ & The constant number    \\
		%	\hline
		%	$\beta$ & The constant number   \\
		%	\hline
			\hline
			$ \theta^{1} $ & A $n \times 4$ matrix that maps from the input layer to the first hidden layer    \\
			\hline
			$ \theta^{j+1} $ & A $n \times 4$ matrix that maps from the $j^{th}$ layer to the $(j+1)^{th}$ layer   \\
			\hline
			$ h_{i}^{(j)} $ & The units' values in layer $j$    \\
			\hline
			$ J\left(\theta\right) $ & The cost function   \\
			\hline
			\hline
	\end{tabular}}
\end{table}

\subsection{System Model}

In Figure~\ref{figures:nrs}, the vehicular network consists of vehicles, edge servers, and base stations. Among them, the base station is responsible for collecting and distributing messages. The edge server performs data processing and transmits the decision results to the vehicle, obtaining the results\cite{azad2018trustvote} by the multi-layer perceptron (MLP) based on Q-learning\cite{na2018soft}. End devices include mobile phones, tablets, laptops, etc., with the assistance of edge servers and base stations, processing data such as reward value, energy consumption, and transmission rate, and combining V2V communication and V2I communication to efficiently and securely transmit the data to the target vehicle.

\subsection{Problem Formulation}
We formulate the data transmission scheduling as the optimization problem in this part, the notations of which can be seen in Table \ref{SN}.

Within the vehicular network, we assume that edge servers, vehicles, base stations, and terminal devices operate jointly to minimize service request time to the greatest extent possible. Considering certain rate-limiting conditions while harnessing the computing resources of vehicles and edge servers, we formulate the scheduling optimization problem in Eq. (1). The first line of the equation delineates the optimization target, which centers on orchestrating the scheduling of data files pertinent to applications with latency-sensitive requirements and allocating these files across a network of base stations. The principal objective of this optimization is to facilitate the reception and return of data near the terminal node, thereby augmenting the capability for data processing while simultaneously minimizing the aggregate cost, subject to constraints imposed by transmission durations.

%The first line is the optimization goal, aiming to schedule the data files of delay-sensitive applications, and distribute the data files among FBSs (Fog-based Base Stations \red{fog base station? what's the context}) so that the overall cost is minimized under the transmission time constraint.
%To optimize the scheduling model including both data transfer and block assignment models, we present the LP formulation, as shown in Eq. \ref{rp}. In Eq. \ref{rp}, line $1$ is the optimization expression, and line $2$ and line $3$ are the limitation. Eq. \ref{rp} aims to seek for scheduling to divide the file blocks of each mobile user's delay-sensitive applications and distribute the file blocks among the FBSs (Fog based Base Stations) along multiple FBS-based network paths, such that the entire cost is minimized.

\begin{align}\label{rp}
\begin{aligned}
& \min \left(P_L \cdot P_A\right)\left(\sum_{n_x \in U} \sum_{E_{x \in i}=A} T T\left(P S_x, U N_{x(I)}\right)\right), \\
& \text { s.t. } \\
& TT\left(P S_x, U N_{x(\varepsilon)}\right) \leq \tau, \forall u_x \in U, f_{x(\varepsilon)} \in A_x \\
& \mathrm{r} \leq \mathrm{r}_0 \\
& \text { Reward } \leq \text { Reward }_{\max } \\
& \mathrm{P}_L \leq T h_L<1 \\
& \mathrm{P}_A \leq T h_A<1 \\
&
\end{aligned}
\end{align}

%\begin{align}\label{rp}
%\begin{aligned}
%\min & \sum_{u_{x} \in U} \sum_{fs_{x(z)} \in A_{z}} TT\left(PS_{x}, UN_{x(z)}\right), \\
%\text { s.t. } & TT\left(PS_{x}, UN_{x(z)}\right) \leq \tau, \forall u_{x} \in %U, fs_{x(z)} \in A_{x}  \\
%& r\leq r_{0}
%\end{aligned}
%\end{align}

$TT\left(PS_{x}, UN_{x(z)}\right)$ denotes the total time required, and $PS_{x}$ is the path set. $UN_{x(z)}$ is the unit of the file block, and $\tau$ is the time. $u_{x}$ is the vehicles of area, and $U$ is the mobile node-set. $fs_{x(z)}$ is the file size, and $A_{x}$ is the mobile application set. $r$ is the transmission rate, and $r_{0}$ is the initial transmission rate. $P_L$ is probability of privacy leakage, and $P_A$ is probability of malicious node attacks. $Th_L$ is the threshold of the probability of privacy leakage, and $Th_A$ is the threshold of the probability of malicious node attacks. $Reward_{max}$ is the maximum reward, reward should be less than the maximum reward. In addition, $TT\left(PS_{x}, UN_{x(z)}\right) = \frac{PS_{x}UN_{x(z)}}{r}$, without considering the priority of data resources, to make data scheduling between vehicles more efficient and secure, it is necessary to reasonably allocate and fully utilize data transmission resources.

To calculate the number of connected vehicles, we use the ratio of ``the number of hello messages received from all one-hop neighbors'' to the ``number of hello messages sent by all one-hop neighbors". Such messages present unique serial number identification and thus can be used to calculate the total messages received. In addition, because a vehicle with a higher antenna has better connectivity to other vehicles, the height of the antenna can be used to measure the number of vehicle connections \cite{wu2020collaborative}, which is calculated as Com (Connectivity Metrics):
%\begin{equation}
%\begin{split}
%C M(x)=\frac{\text { bandwidth . velocity of receiving vehicle}}{\text { velocity of sending vehicle }}. \\ 
%\frac{\text { transmission rate of receiving vehicle}}{\text { transmission rate of sending vehicle }},
%\end{split}
%\end{equation}
%Note that ``the number of hello messages sent by all one-hop neighbors" can be calculated by observing the sequence number of received hello messages since each hello message is identified with a unique sequence number which is incremented by a predefined value for each hello period. The other way is to use the antenna height to show the connectivity metric as a vehicle with a higher antenna always can provide better connectivity to the neighbor vehicles. In that case, Com (Connectivity Metrics) is calculated as:
\begin{equation}
Com(x)=\frac{B_{x} \cdot V_{r} \cdot r_{r}}{V_{s} \cdot r_{s}}
\end{equation}
where $B_{x}$ is the bandwidth of node $x$, $V_{r}$ is the velocity of vehicles that receive message, $r_{r}$ is the data transmission rate, $V_{s}$ is the velocity of sending vehicle, and $r_{s}$ is the transmission rate of sending vehicle. The state is a pair of the current node and the destination node, and the action is the selection of the next-hop node. In V2I communication, the next hop is the base station; in V2V communication, the next hop is the neighboring vehicle. 

%\red{confused to see how the q learning is useful in here, only state is mentioned above, so context is required. }The Q-learning model is introduced here. 

	\section{Data Scheduling based on Q-Learning (DSQL)} \label{section:algorithm}

To elevate the probability of data transmission, we propose Q-learning-empowered data scheduling in this section, including Q-learning-based data communication, multi-layer perceptron-based prompt data processing, and differential privacy-based security preservation.

% It is an adaptive system used to model the complex relationship between input and output or to explore data patterns. Neural networks \cite{9335241} are now widely used to solve problems in control, pattern recognition, and data analysis. 
% %The functions of the neural network include high processing speed and the ability to learn problem solutions from sample examples.

% An algorithm of data scheduling based on Q-learning is proposed in the vehicular network, named DSQL. The DSQL guarantees data transmission by Q-learning and the information is quickly processed by the shallow perceptron network. The algorithm plays a crucial role in the intermittently connected network. The information is processed quickly and autonomously to schedule vehicles. The DSQL mainly considers power consumption, energy consumption, and rewards. Q-learning-based data communication is applied to guarantee data transmission. The information is processed quickly by the way of neural network-based data processing, and the transmission delay of the data is optimized in a 5G network.

\begin{figure}
	\centering
	\includegraphics[width=0.9\linewidth]{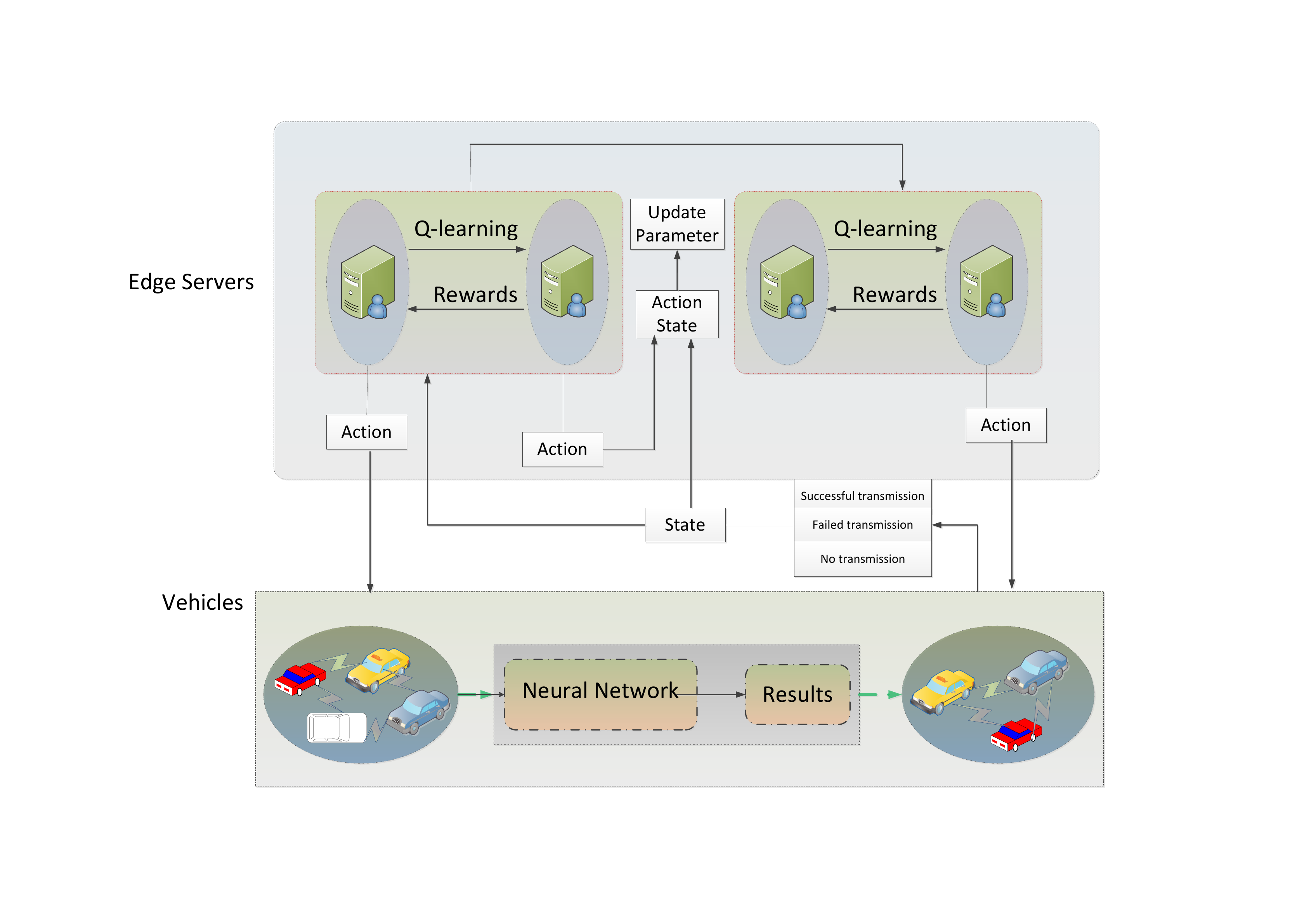}
	\centering{\caption{An Overview of the proposed DSQL algorithm.}\label{figures:overview}}	 	
\end{figure}

\subsection{Overview of DSQL}

Q-learning, classified as a model-free reinforcement learning algorithm, has been empirically validated for its robust performance in vehicular environments that are both dynamic and uncertain. It demonstrates notable flexibility in adapting to environmental variations. Furthermore, it offers the benefit of supporting offline training regimens, utilizing datasets that have been pre-deployed. Q-learning determines the optimal decision-making process by establishing relationships among states, actions, and reward values. Specifically, status represents various data transmission statuses, including success, failure, or not transmission. When data transmission is successful, the next hop node receives the reward. Through Q-learning training, the edge server obtains data transmission decisions to adjust the actions of vehicle nodes and edge server nodes.

DSQL consists of three parts, Q-learning empowered data communication, data processing based on multi-layer perceptron, and data privacy protection (as shown in Figure \ref{figures:overview}). In data communication, Q-learning is performed by exchanging results and parameters after acquiring data. First, calculate the power consumption and energy consumption based on the CPU frequency. Secondly, the amount of data is calculated based on the time difference and CPU frequency. Finally, rewards are calculated based on latency, packet size, and bandwidth. In data processing based on a multi-layer perceptron, the multi-layer perceptron includes six layers, namely the input layer, four hidden layers, and the output layer. Obtain data based on multilayer perceptron processing. The state in Q-learning is used as the input information of the multi-layer perceptron, and the income of the action, that is, the Q value, is used as the output information of the multi-layer perceptron. Finally, in data privacy protection, differential privacy and interference conditions are used to protect private information.

First, the reward value, energy consumption, and transmission rate are calculated in Q-learning. Then the multi-layer perceptron is used to process the data quickly, and vehicles with high reward value, low energy consumption, and fast transmission rate are selected to make optimal decisions for data transmission scheduling. Q-learning updates the Q value and calculates the corresponding reward value under different conditions, as shown in Formula (4) (5) (6) (7), and then calculates the energy consumption value based on the CPU frequency and distance. Finally, calculate the transmission rate, and combine these three factors as the input of the multi-layer perceptron for data processing. During the data processing process of the multi-layer perceptron, an additional mapping process of the sine function is performed in the activation function so that the value does not exceed 1, thus reducing the difficulty of calculation.
%\red{what's the connection between q learning and vanilla neural network}

In the following, we first describe the Q-learning-based data communication and then the multi-layer perceptron-based data processing. Finally, the amount of data transmitted is analyzed to elevate efficiency, and data privacy is protected by differential privacy. The pseudo-code of the DSQL is shown in Algorithm \ref{aone}. Inputs are the amount of CPU frequency $f_{n}^{\text {local }}$ and the bandwidth $ B $. Outputs are energy consumption $E_{n}^{\text {local }}$, the amount of processing data $D_{n}^{\text {local }}$, and transmission rate $r_{n, k}^{t, V2I }$.
%\begin{algorithm}[!ht]
%	\caption{Data Scheduling based on Q-learning (DSQL)}
%	\label{alg:Framwork}
%	\begin{algorithmic}[1]
%		\REQUIRE $f_{n}^{\text {local }}$, $ p_{n}^{\text {local }} $, $B$, $Com_{x}$
%		\ENSURE $E_{n}^{\text {local }}$, $D_{n}^{\text {local }}$, $r_{n, k}^{t, V2I }$
%		\STATE Calculate the energy of vehicles in the dense area;
%		\STATE Calculate the amount of processing data in the dense areas;
%		\IF {The energy of vehicles is in a certain range}
%		\STATE The information is processed by the enhanced multi-layer perceptron;
%		\ELSE
%		\STATE $p_{n}^{\text {local }}=\frac{B\left(f_{n}^{\text {local }}\right)^{2}}{d}$;
%		\ENDIF
%		\WHILE {The reward value is the minimum}
%		\STATE  $E_{n}^{\text {local }}=p_{n}^{\text {local }} \Delta t=\frac{B\left(f_{n}^{\text {local }}\right)^{2} \Delta t}{d}$ ;
%	    \STATE Use distance $d$ to compute the vehicle's amount of data;
%		\STATE  $D_{n}^{\text {local }}=\frac{f_{n}^{\text {local }} \Delta t}{d \cdot c}$;
%		\STATE Ensure data communication through the amount of processing data $D_{n}^{\text {local }}$;
%		\ENDWHILE
%		\STATE Achieve the constraints by the computing of edge servers;
%        \STATE Achieve data privacy protection by differential privacy;
%		\STATE Obtain the minimization of the energy;
%		\label{code:fram:select6}
%	\end{algorithmic}\label{aone}
%\end{algorithm}
\begin{algorithm}[!ht]
	\caption{Data Scheduling based on Q-learning (DSQL)}
	\label{alg:Framwork}
	\begin{algorithmic}[1]
		\REQUIRE $f_{n}^{\text {local }}$, $B$
		\ENSURE $E_{n}^{\text {local }}$, $D_{n}^{\text {local }}$, $r_{n, k}^{t, V2I }$
      \WHILE {Receiving rewards through V2I}
		\STATE $Red=
\bar{Red} \cdot D_{n}^{local}$ 
		\ENDWHILE
\WHILE {Receiving rewards through V2V}
		\STATE $Red=
D_{n}^{local} $ 
		\ENDWHILE
\WHILE {Receiving rewards through the cellular interface}
		\STATE $\hat{Red}=\min \left(1, \frac{PD_{t h}+D_{n}^{local} \cdot T_{persistent}}{CB+PD_{b s}}\right)$ 
		\ENDWHILE
\WHILE {Receiving rewards through the IEEE 802.11p communication protocol}
		\STATE $\hat{Red}=\min \left(1, \frac{PD_{t h}\cdot T_{persistent}}{\frac{PS}{CB_{11 p} \cdot D_{n}^{local}\cdot HRR }\cdot P_{obstacle}+PD_{11p}}\right)$ 
		\ENDWHILE
		\IF {The energy of vehicles is in a certain range}
\STATE Calculate the energy of vehicles in the dense area;
\STATE $p_{n}^{\text {local }}=\frac{B\left(f_{n}^{\text {local }}\right)^{2}}{d}$;
\STATE  $E_{n}^{\text {local }}=p_{n}^{\text {local }} \Delta t=\frac{B\left(f_{n}^{\text {local }}\right)^{2} \Delta t}{d}$ ;
		\STATE The information is processed by the enhanced multi-layer perceptron;
\STATE Calculate the amount of processing data in the dense areas;
\STATE Use distance $d$ to compute the vehicle's amount of data;
		\STATE  $D_{n}^{\text {local }}=\frac{f_{n}^{\text {local }} \Delta t}{d \cdot ad}$;
		\ELSE
		\STATE Ensure data communication through the amount of processing data $D_{n}^{\text {local }}$;
\STATE Obtain the minimization of the energy;
		\ENDIF
\WHILE {Data are offloaded from vehicle $n$ to RSU $k$ on a licensed V2I channel}
		\STATE $r_{n, k}^{t, V2I }=B \log _{2}\left(1+\frac{p_{n}^{\mathrm{tr}}|h|^{2}}{\omega_{0}\left(d_{n, k}^{t}\right)^{\vartheta}}\right)$ 
		\ENDWHILE
		\STATE Achieve the constraints by the computing of edge servers;
        \STATE Achieve data privacy protection by differential privacy;
		\label{code:fram:select6}
	\end{algorithmic}\label{aone}
\end{algorithm}

\subsection{Q-Leanrning Based Data Communication}

By employing reinforcement learning principles, Q-Learning facilitates iterative learning and enhancement through rewarding correct operations and penalizing incorrect ones, ultimately enabling the discovery of the shortest path for message communication. Q table is used to maintain the current information and historical information throughout the decision-making in Q-learning.
%\red{again, lost of context, q table and definition of q learning should be explained before using. }
The rows represent different states, and the columns represent the actions in the corresponding states. Through the interaction between the agent and the environment, the reward or punishment is fed back, and the value of the Q table is updated iteratively to reflect the learning content.
%\red{to my understanding, the q learning is used to find the shortest path of message communication?} 
The Q value is determined by the triplet of the destination node, the communication type, and neighboring vehicles, and the Q table is updated after receiving the hello message. The initial value of the Q value is set to $0$, the space of the Q table is released as the vehicle moves, and the information of the new neighbor is stored. To save the space of the Q table, the vehicle as an edge node uses hierarchical routing to find the path. When the node $m$ receives the Hello message, it will update the Q value accordingly:
%The state is expressed by a pair of {destination, current node}, and the action is determined by {communication type, next hop}. In the case of using cellular communications, the next-hop would be BS, and in the case of IEEE 802.11p, the next hop node would be a neighbor node. Each node has to maintain a Q-value for each triple of a destination, the communication type, and a one-hop neighbor. Upon reception of each hello message, the Q-Table is updated. The initial value for each Q-value is 0. Each vehicle maintains a Q-value to the BS and each vehicle in a two-hop region. Considering the change of neighbors with the vehicle movement, we release the corresponding Q-Table space of old neighbors when necessary for maintaining information about new neighbors. The proposed scheme does not maintain a route for each possible destination considering the size of the Q-Table. For finding a route to other nodes, the proposed scheme uses a hierarchical routing approach where different levels of gateway nodes exist. Note that for each vehicle, there is at least one neighbor would be an edge node that is working as a gateway node and responsible for finding a route to any other vehicle. The BS also performs the duty of gateway. After reception of a hello message from node $m$, node $c$ updates the corresponding Q-value as
\begin{equation}\label{aa}
\begin{aligned}
Q_{c}(dn, te, m) \leftarrow & \alpha \times L Q(c, m) \\
& \times\left\{Red+\beta \times \max _{y \in N B_{m}} Q_{m}(dn, te, y)\right\} \\
&+(1-\alpha) \times Q_{c}(dn, te, m)
\end{aligned}
\end{equation}
where $dn$ and $te$ represent the destination node and communication type, respectively. The link quality value $LQ(c, m)$ of the vehicle $c$ and the vehicle $m$ is expressed through the received message ratio between the two vehicles, and the single-hop neighbor set of the node $m$ is $N B_{m} $.

When receiving rewards through V2I, the reward value $Red_{m}$ is
\begin{equation}\label{ref1}
Red=\left\{\begin{array}{ll}
\bar{Red} \cdot D_{n}^{local}, & \text { if } m \text { is base station} 
%c \text { is an edge node} 
\\
0, & \text { otherwise }
\end{array}\right.
\end{equation}
Among them, the BS distributes the reward value $\bar{Red} \in[0,1]$ according to the number of connected vehicles. BS sets the reward value according to the density of the vehicle, and the higher the density, the greater the reward value. But the farther the number of hops, the smaller the reward value. The reward value is explored through the exchange of hello messages, so the node path is selected for communication through the reward value.

When receiving rewards through V2V communication, the reward value $Red$ is
\begin{equation}\label{ref2}
Red=\left\{\begin{array}{ll}
D_{n}^{local}, & \text { if } m \text { is an edge node } \\
0, & \text { otherwise }
\end{array}\right.
\end{equation}

To utilize distributed communication, the reward value $Red$ is set to $D_{n}^{local}$. Distributed communication methods include IEEE 802.11p and millimeter waves. Because millimeter waves only exist in the line of sight communication link, they are used for edge vehicles. When both are available, a millimeter wave is selected.

Calculate the reward value through the cellular interface:
\begin{equation}\label{rrp}
\hat{Red}=\min \left(1, \frac{PD_{t h}+D_{n}^{local} \cdot T_{persistent}}{CB+PD_{b s}}\right)
\end{equation}
where $PD_{th}$, $T_{persistent}$, $CB_{ul}$, $CB_{dl}$, $PD_{bs}$ are the delay requirements, the duration of the vehicle's movement, the uplink bandwidth, the downlink bandwidth, and the base station processing delay. According to the processing capacity of the BS, the reward value is set to a value less than or equal to $1$. Processing delay includes transmission, scheduling, queuing, and calculation time.

Upon receiving a reward through the IEEE 802.11p communication protocol, the reward can be derived as:
\begin{equation}\label{rrpp}
\hat{Red}=\min \left(1, \frac{PD_{t h}\cdot T_{persistent}}{\frac{PS}{CB_{11 p} \cdot D_{n}^{local}\cdot HRR }\cdot P_{obstacle}+PD_{11p}}\right)
\end{equation}

$CB_{11p}$ is the bandwidth of the communication protocol, $PS$ is the number of data packets, and $HRR$ is the reception ratio of hello messages between two vehicles. $PD_{11p}$ is the processing delay of the vehicle under this communication protocol, including contention delay and re-transmission delay. To avoid transmission that is more than two hops, the discount rate is set to $0.5$. The training process is 1. Initialize the Q table, Q(s, a) = 0, or a small random value. 2. Cycle training: Select action a based on the current state and exploration rate $\epsilon$. If the random number $<$  $\epsilon$, select a random action (exploration). Otherwise, select the action with the largest Q value (exploitation). Execute action: Execute action a in the environment and observe the resulting state s' and reward r. Update Q value: Update Q value according to the following formula. Update state: Update state s to the new state s'. 3. Repeat the above steps until the Q value exceeds the threshold. The parameters settings are 1. Learning rate ($\alpha$): The learning rate determines the weight of the old and new values each time the Q value is updated, ranging from 0 to 1. A higher learning rate can quickly learn new information but may cause shocks. A lower learning rate makes learning more stable, but slower; 2. Discount factor ($\gamma$): The discount factor is used to balance the weight of immediate rewards and future rewards, ranging from 0 to 1. A higher discount factor will pay more attention to long-term rewards, and a lower discount factor will pay more attention to immediate rewards; 3. Exploration rate ($\epsilon$): The exploration rate determines the probability of selecting a random action during training, ranging from 0 to 1. As training progresses, the exploration rate usually decreases, so that more optimal actions are selected; 4. Action-state value (Q value): The Q value represents the expected total reward for taking an action in a state. The parameters Settings are as Table \ref{PTSa}.

\begin{table}
	\caption{Q-learning Parameter Settings}\label{PTSa}\centering
	\scalebox{1.0}{
		\begin{tabular}{l|c}
			\hline
			\hline
			Learning rate & $0.5$    \\
			\hline
			Diccount rate & $0.5$    \\
			\hline
			Exploration rate & $0.7$     \\
			\hline
			Action-State Value (Q Value) & $ 8$   \\
			\hline
			\hline
	\end{tabular}}
\end{table}

\begin{figure}
	\centering
	\includegraphics[width=0.5\textwidth]{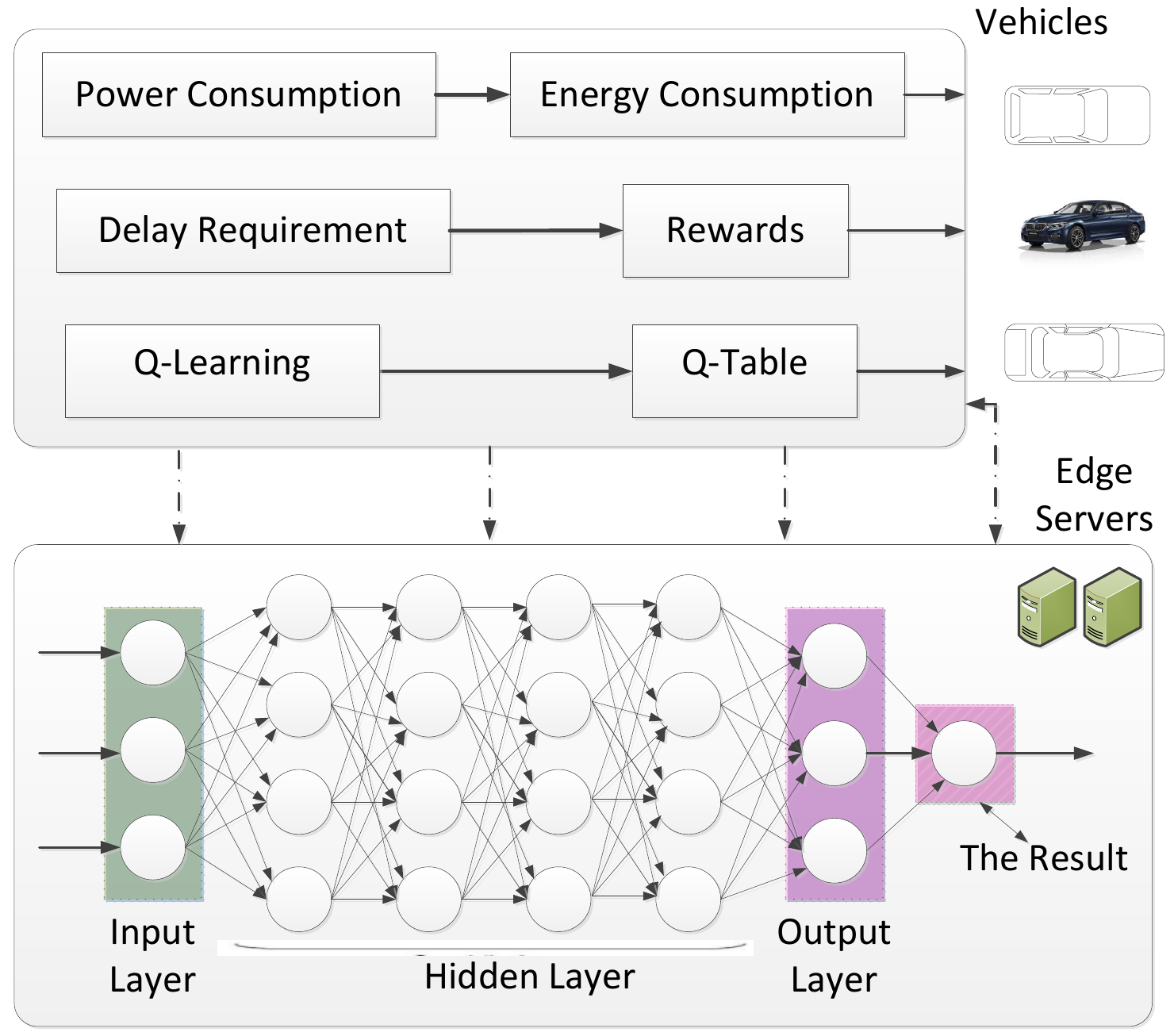}
	\centering{\caption{{Q-learning-based data computation.}}\label{figures:neN}}
	\vspace{-3mm}	
\end{figure}

In Q-learning-based data communication, as shown in Figure \ref{figures:neN}, data is transmitted for lower energy. The energy consumption is obtained through distance, transmission power, and the amount of CPU frequency. 

The processing capacity of the vehicle $n$ for local calculation is $f_{n}^{\text {local }}$ (that is, the frequency of the CPU, in cycles per second), and the power of the vehicle $n$ for processing local data is
\begin{equation}
p_{n}^{\text {local }}=\frac{B\left(f_{n}^{\text {local }}\right)^{2}}{d}
\end{equation}
The channel bandwidth of the vehicle $n$ is $B$. So in a time slot, the energy consumption of the vehicle $n$ to process local data is
\begin{equation}
E_{n}^{\text {local }}=p_{n}^{\text {local }} \Delta t=\frac{B\left(f_{n}^{\text {local }}\right)^{2} \Delta t}{d}
\end{equation}
In a certain period, the amount of data processed locally by the vehicle $n$ is
\begin{equation}
D_{n}^{\text {local }}=\frac{f_{n}^{\text {local }} \Delta t}{d\cdot ad}
\end{equation}
where $ad$ is the amount of data processed per unit of time.

In the vehicular network, the data can be offloaded to RSU through V2I communication scheduled by the control center. To model the data offloading through V2I communication, we denote the path loss as $d^{-\vartheta}$, where $d$ and $\vartheta$ denote the distance from the transmitter to the receiver and the path-loss exponent, respectively. Moreover, the channel fading coefficient is denoted by $h$, which is modeled as a circularly symmetric complex Gaussian random variable. When data are offloaded from vehicle $n$ to RSU $k$ on a licensed V2I channel, the transmission rate is given by:
\begin{equation}\label{bb}
r_{n, k}^{t, V2I }=B \log _{2}\left(1+\frac{p_{n}^{\mathrm{tr}}|h|^{2}}{\omega_{0}\left(d_{n, k}^{t}\right)^{\vartheta}}\right)
\end{equation}

In Eq. (11), $p_{n}^{\mathrm{tr}}$ is the transmission power, and $\omega_{0}$ is the Gaussian white noise power. The distance between the vehicle $n$ and RSU $k$ during the time $t$ is $d_{n,k}^{t}$. The energy consumption of data transmission is $E_{n}^{\mathrm{tr}}=p_{n}^{\mathrm{tr}} \Delta t$.

Data communication based on Q-learning is a data communication method based on reinforcement learning. The behavior and decision-making of the data communication system are learned and optimized by the Q-learning algorithm.
Q-learning is a reinforcement learning algorithm used for problem decision-making. Q-learning optimizes the decision-making of communication systems under different states and actions to improve communication performance. The data communication process based on Q-learning is as follows:
(1) First, define the status of the data communication system, including data transmission success, failure, and no transmission; (2) Define the action of the communication system, that is, the selection of the next hop node; (3) Compute the reward values obtained for different actions; (4) Q-Table initialization: Create Q-Table to store the Q value corresponding to the state and action, that is, the long-term return of the action in a certain state; (5) Training process: Update the value of Q-Table based on the results of interaction with the environment; (6) Decision-making process: After training is completed, select the action with the highest Q value, which is the current decision.

The training and decision-making process of the Q-learning algorithm enables the data communication system to select optimal actions, adapt to different communication environments and conditions, improve communication efficiency, and achieve the purpose of optimizing network performance.

\subsection{Data Processing Based on Multi-layer Perceptron}

The edge server utilizes a shallow multi-layer perceptron network to perform rapid processing after the data is transmitted. To prevent the obtained results from being attacked by malicious nodes, the method of Q-learning is adopted for the data communication of nodes. A multi-layer perceptron network essentially serves as a mapping function from input to output and implicitly depends on trainable model parameters\cite{lin2021mednet} as shown in Figure ~\ref{figures:neN}. 

\begin{figure}
	\centering
	\includegraphics[width=0.5\textwidth]{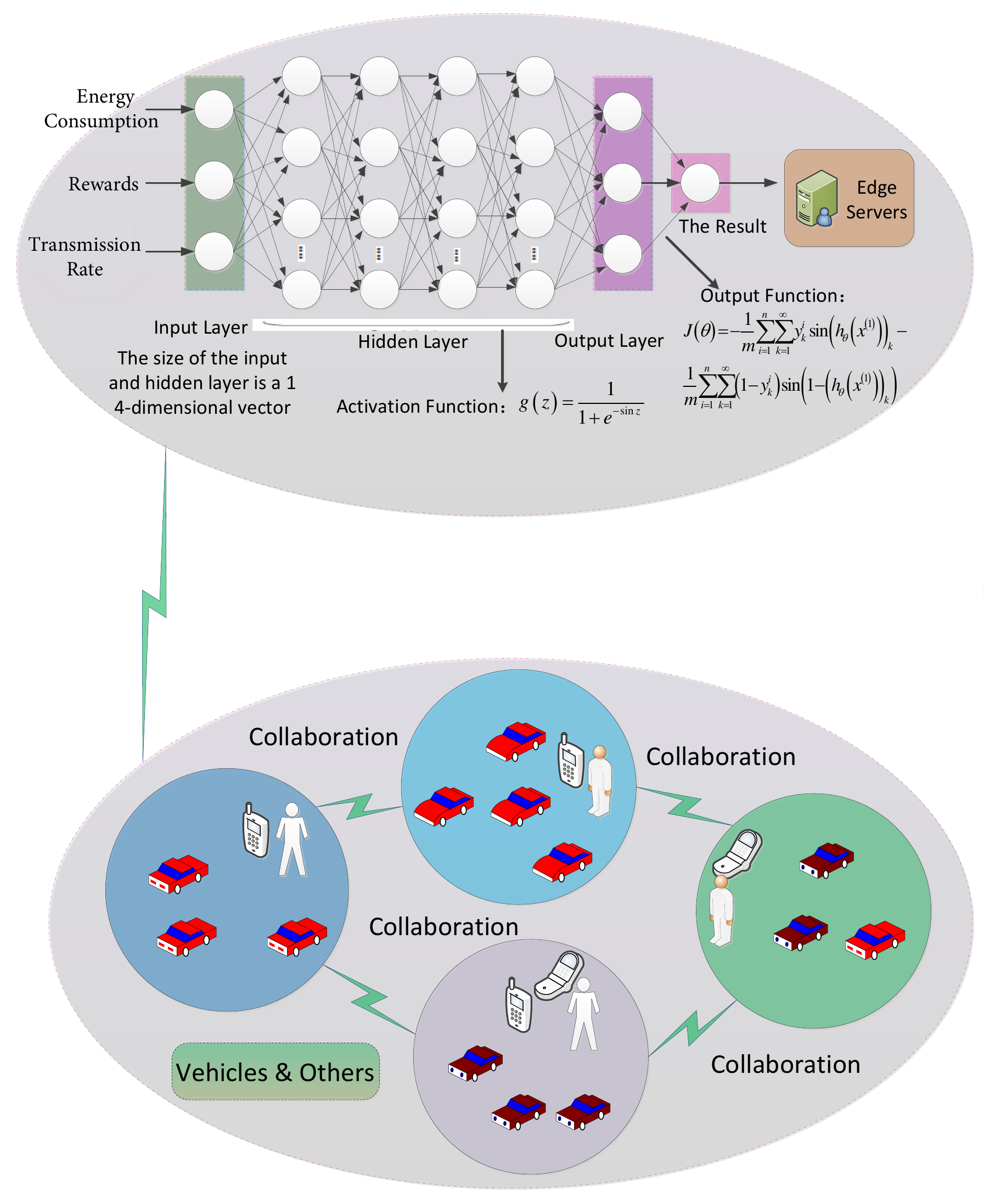}
	\centering{\caption{{Multi-layer perceptron model.}}\label{figures:ne}}
	\vspace{-3mm}	
\end{figure}

The network is shown in Figure~\ref{figures:ne}. It includes the input layer, output layer, and hidden layers. Inputs to the multi-layer perceptron are reward values (derived from Eq.
(4), (5), (6), and (7) are subject to different communication methods), transmission rates, and energy consumption, respectively. Output is a communication decision, that is, selecting the next-hop optimal node for transmitting data. It aims to reduce the scope of data processing of the vehicular network and classify the data in a small range. Due to the characteristics of the rapid movement of the vehicle, we analyze and process the data of the vehicles to save as much time as possible. The decision of the vehicle is assisted to be made, and the task of data transmission is quickly completed. The training process is 1. Data preparation: Divide the data into training sets and test sets. Perform standardization or normalization to make the data within the appropriate range. 2. Model initialization: Define the network structure of MLP, including the number of network layers and activation function. 3. Model training: Forward propagation: The input data is propagated through the network layer by layer to obtain the predicted output. Calculate the loss: Use the loss function to calculate the error between the predicted output and the true value. Backward propagation: Calculate the gradient of the loss relative to each parameter. Parameter update: Use the optimizer to update the network parameters according to the gradient. Repeat the above steps: Train in batches until all iterations are completed. 4. Model evaluation: Use the test set to evaluate the model performance, such as accuracy. The parameters settings are 1. Network structure parameters: The number of layers includes the input layer, one or more hidden layers, and the output layer. Common activation functions include ReLu, Sigmoid, and Tanh. Different activation functions can be used for each layer; 2. Learning rate: The learning rate determines the step size of each parameter update, usually ranging from 0 to 1. A higher learning rate can speed up training but may cause instability. With a lower learning rate training is more stable but slower; 3. Optimizer: The optimizer determines how to update the network weights. Common optimizers include SGD (stochastic gradient descent), Adam, RMSprop, etc. Adam is usually the default choice because it performs well in many cases; 4. Loss function: The loss function is used to measure the gap between the model prediction and the true value. Common loss functions include mean square error (MSE) for regression and cross-entropy loss (Cross-Entropy Loss) for classification; 5. Batch Size: Batch size determines how many training samples are used for each parameter update. Larger batch sizes can increase training speed but require more memory. Smaller batch sizes can make training more stable; 6. Epochs: Epochs refers to the number of times the entire training dataset is fully fed into the multilayer perceptron for training. The parameters Settings are as Table \ref{PTSb}.

\begin{table}
	\caption{Multi Layer Perceptron Parameter Settings}\label{PTSb}\centering
	\scalebox{1.0}{
		\begin{tabular}{l|c}
			\hline
			\hline
			Network Structure & Input layer, four hidden layers, \\
            %\hline
               & and output layer, activation function \\
            %\hline
               & is Formula (12)  \\
			\hline
			Learning rate & $0.5$    \\
			\hline
			Optimizer & $Adam$     \\
			\hline
			Loss function & Formula (14)   \\
	\hline
			Batch size & $180$   \\
	\hline
			Epochs & $12$   \\
			\hline
			\hline
	\end{tabular}}
\end{table}

%Inputs are distance $l_{ab}$, the amount of CPU frequency $ s_{i} $, the channel fading coefficient $ C_{i,j} $, and packet size $r_{i,j}$. For example, if the distance is $20$ meters, the amount of CPU frequency is $30$, the channel fading coefficient is $202$, and the paket size is $1022$ bit. The input matrix is four-dimensional vector, and these values are moderate. Outputs are energy consumption $H$, the amount of data $\rho(A)$, and rewards $\left(f_{br}(\xi,\mu)\right)$. For example, if the energy consumption is $32.2$, the amount of data is $5252$ bit, and the value of the rewards is $25$. The output matrix is three-dimensional vector, and the efficiency and privacy of data transmission is better according to the results of output. 

$g(.)$ is an activation function in every neuron for hidden layers, where
\begin{equation}
g(x)=\frac{1}{1+e^{-sin(x)}}
\end{equation}

\begin{figure}[t]
	\centering
	\subfigure[\label{figures:nplrf}
	Activation Function.] 
	{\includegraphics[width=2.6in,angle=0]{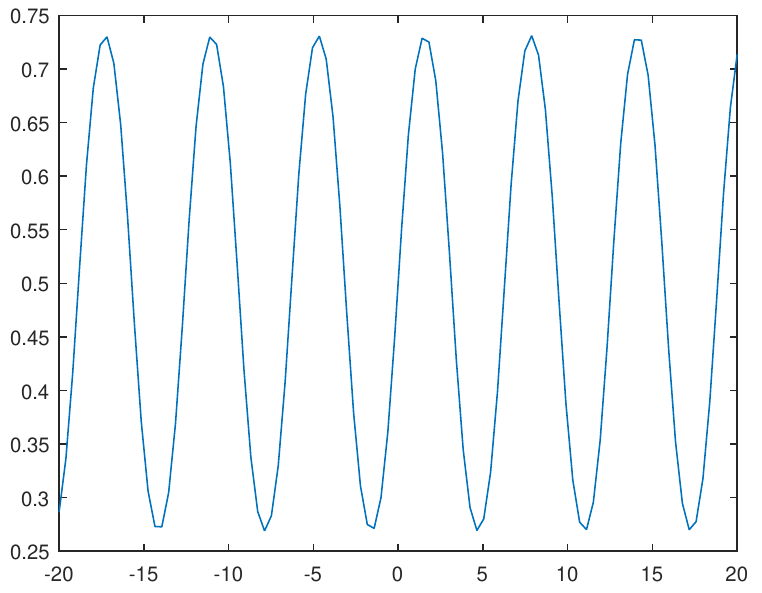}}
	\subfigure[\label{figures:nplrg}
	Sigmoid Function.] 
	{\includegraphics[width=2.6in,angle=0]{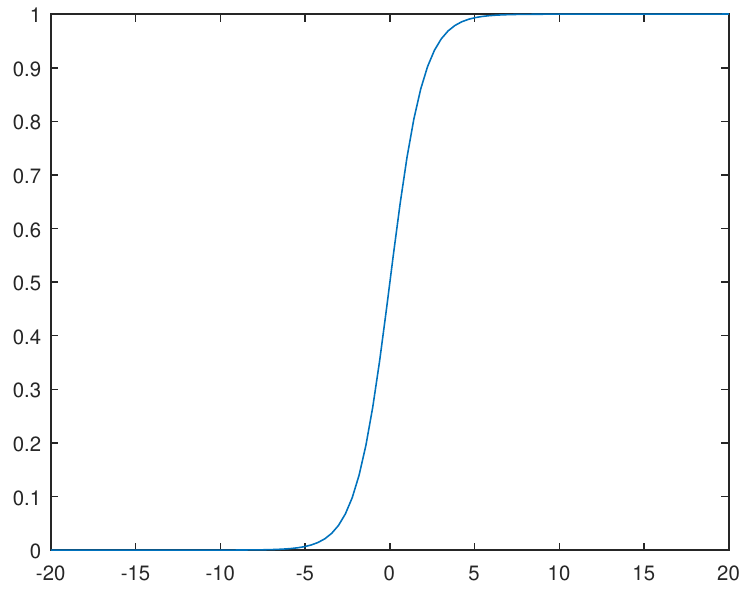}}
	\caption{Activation Function for Different Data Processing Range.}
\end{figure}

As shown in Figure \ref{figures:nplrf} and Figure \ref{figures:nplrg}, the pre-defined data processing range of the activation function is from $0.25$ to $0.75$, and the range of the sigmoid function is $0$ to $1$ under the same data. The rationale is that the smaller the data processing range of the activation function, the higher the data processing efficiency.

The hypothesis function can be obtained from the activation function:
\begin{equation}
h_{\theta}(x)=g\left(\theta^{T} x\right)
\end{equation}

We use backpropagation (BP) to minimize the cost function. To make the deviation between the obtained data and the true value smaller, the cost function is designed as follows according to the hypothesis function:
\begin{multline}
J\left(\theta \right)  
=-\frac{{1}}{m} \sum_{k_{1}=1}^{n}\sum_{k_{2}=1}^{\propto}y_{k_{2}}^{k_{1}}sin \left(h_{\theta}\left(x^{\left(1\right)}\right)\right)_{k_{2}}  \\
-\frac{{1}}{m} \sum_{k_{1}=1}^{n}\sum_{k_{2}=1}^{\propto}\left(1-y_{k_{2}}^{k_{1}}\right)sin\left(1-\left(h_{\theta}\left(x^{\left(1 \right)}\right)\right)_{k_{2}}\right) 
\end{multline}
where $m$ is the amount of training set samples. The label $y_{k_{2}}^{k_{1}}=1$ if the $k_{1}^{th}$ sample belongs to the $k_{2}^{th}$ unit type, otherwise $y_{k_{2}}^{k_{1}}=0$. $k_{1}$ represents a sample, and $k_{2}$ represents a unit type. $x^{(k{_{1}})}$ represents the $k_{1}^{th}$ training set sample, and $\left(h_{\mathrm{\theta}}\left(x^{(1)}\right)\right)_{k_{2}}$ is the $k_{2}^{th}$ output neuron's value mapped from input $x^{(k_{1})}$. $\left(h_{\mathrm{\theta}}\left(x^{(1)}\right)\right)_{k_{2}}$ is the hypothesis function of $\theta$. When $y_{k_{2}}^{k_{1}}=1$, the first term works, the larger $\left(h_{\mathrm{\theta}}\left(x^{(1)}\right)\right)_{k_{2}}$ is, the better, which means it is closer to the true value; when $y_{k_{2}}^{k_{1}}=0$, the second term works, the smaller the $\left(h_{\mathrm{\theta}}\left(x^{(1)}\right)\right)_{k_{2}}$, the better, which means it is closer to the true value. We find an optimal $\theta$ through the perceptron network by minimizing the cost function $J(\theta)$. Next, we analyze the aspects of gradient quantization and sparsification, gradient compression, and data compression:

Due to the high cost of data storage and transmission in machine learning models, quantification can be used to compress big data to reduce costs, and also reduce the cost of communication, so that the model has better convergence. In all gradient data, only valuable gradient data is sent to achieve the purpose of gradient sparse. Gradient values that meet certain conditions are reserved, and other gradient values are uniformly processed to perform sparse update operations.

To predict the communication time, the communication process of distributed training is controlled. Adjusting the compression level according to the dynamic characteristics of the network does not require unification of compression during the entire training process, and does not affect the convergence of the model. Because the compression level cannot be known in advance, it is difficult to use compression to speed up training throughout the training process. Weigh the relationship between training time and accuracy based on data compression and network conditions. To relieve network congestion, different levels of compression are applied when needed. Through the trade-off between time and accuracy, when the network conditions are stable and not congested, the accuracy of the model is ensured, and the training of the model is completed by using compression. Compression control through adjustment of gradient compression levels, making it a lightweight proxy communication library. And integrate it into the software to realize gradient compression control in model training.

Data compression is closely related to the probability distribution. Lossless compression of data is limited by the entropy of probability distribution. Therefore, data compression can be understood as a probability model of sequence data. The better the probabilistic model is learned, the better the compression. Compression algorithms are improved by learning Bayesian nonparametric models, progressing with the development of probabilistic machine learning. Data compression is the process of converting more byte sequences into fewer byte sequences while containing the same information. Data compression is divided into bit compression and byte compression. Bit compression has a higher compression ratio, and byte compression has high throughput, no need for shifting operations, less hardware consumption, and is suitable for partitioning and parallel processing.
\subsection{Data Privacy Protection}

%\red{which data's privacy do we want to protect in here, reward? or other?}
In the data transmission of the vehicle, the data privacy of reward value and action of the vehicle is protected by the differential privacy mechanism. Within the available reward value range, set the system random parameters of differential privacy so that they meet the interference conditions shown in inequality \ref{figures:aap}.

\begin{figure}
	\centering
	\includegraphics[width=0.4\textwidth]{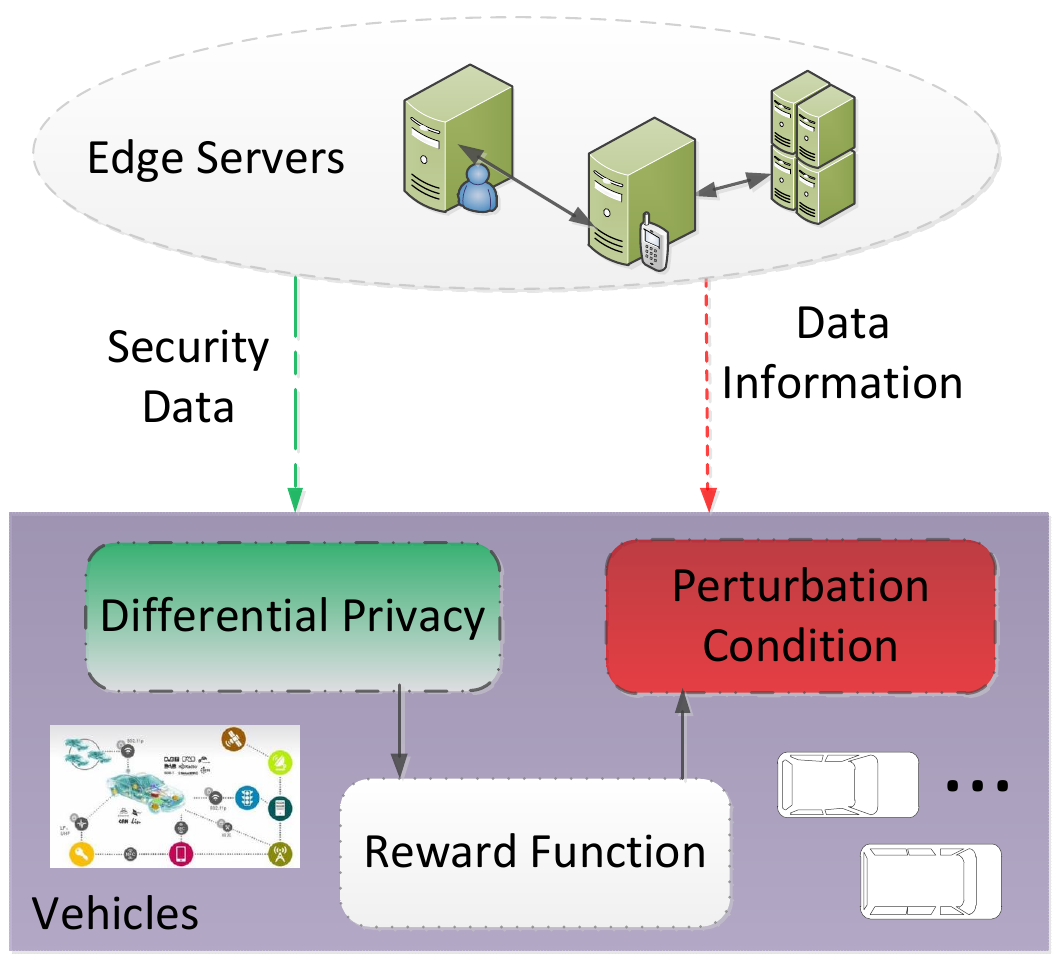}
	\centering{\caption{{Data privacy protection.}}\label{figures:ddp}}
	\vspace{-3mm}	
\end{figure}

As shown in Figure \ref{figures:ddp}, by satisfying the interference conditions and at the same time ensuring that the reward value is maximized, the edge server can obtain secure data results. Apply differential privacy for data privacy protection. First, differential privacy is explained as follows: $M :(\chi \cup\{\perp\})^{N \times K} \rightarrow R^{K \times 1}$ indicates that the input data matrix is mapped to synthetic interference vector $R$. $(\chi \cup\{\perp\})^{N\times K}$ represents the data of all users, $N$ represents a group of users, and $K$ represents a group of communication data. Then, $M$ is $\eta$-difference Privacy equivalent to two different data matrices $x$ and $x'$ in only one entry and $A \subseteq R^{ K \times 1}$, So there is the following inequality:
\begin{equation}
\operatorname{Pr}[M(x) \in A] \leq \exp (\eta) \operatorname{Pr}\left[M(x') \in A\right]
\end{equation}
where $\eta$ is the privacy budget, usually a small positive number. When protecting information privacy, the size of the reward value needs to be considered. The reward value is related to the privacy budget set by the system.

Since pseudonym entropy plays the role of protecting vehicle data, pseudonym entropy is introduced to strengthen the protection of vehicle communication data. When the pseudonym entropy meets certain conditions, it indicates that the information has a certain security basis. Otherwise, the communication data is not considered for further protection.

The pseudonym entropy of a vehicle is defined as $H=-\sum_{i=1}^{k} p_{i} \log _{2} p_{i}$. In particular, pseudonym entropy decreases with increasing probability. $p$ is the probability of being attacked by a malicious node. $H=\left[H_{i, j}\right] \in[0,1]^{N \times K}$ represents the user's pseudonym entropy matrix. Assuming that the pseudonym entropy matrix is prior knowledge, the edge server will keep a historical record of the pseudonym entropy matrix, which is obtained by the probability of being attacked by malicious nodes.

\begin{figure}[ht]
	\centering
	\includegraphics[width=0.5\textwidth]{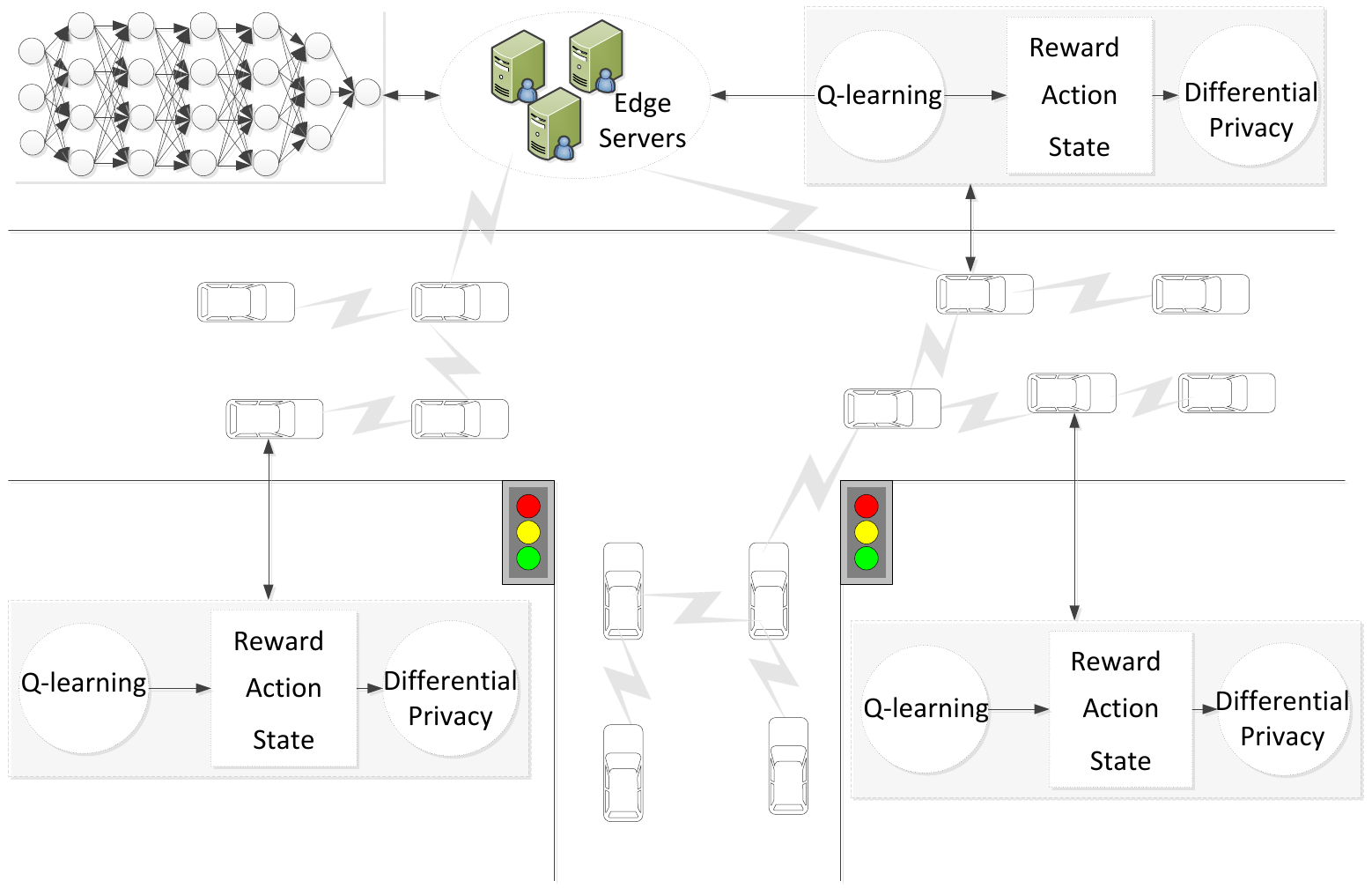}
	\centering{\caption{Practical vehicular network.}\label{figures:pvN}}
	%\vspace{-3mm}	
\end{figure}

In Q-learning, when a traffic-intensive application is performed, the reward value is calculated as $Red$, that is, Eq. (4) and (5); when a delay-aware unicast application is performed, the reward value is calculated as $\hat{Red}$, that is, Eq. (6) and (7); this is expressed in the following equation:
\begin{align}
\operatorname{Reward}=\begin{cases} Red & \text { Traffic-intensive Applications } \\ \hat{Red} & \text { Delay-sensitive Unicast Applications }\end{cases}
\end{align}
Interference conditions need to meet the requirements shown below:
\begin{equation}
Pr\left[\hat{X}_j \neq x_j^*\right] \leq \frac{\lambda_j+H}{4}\label{figures:aap}
\end{equation}
where $\hat{X}_{j}$ is a random variable, $x_{j}^{*}$ is the real data, H is the pseudonym entropy, and $\lambda_j$ is the parameter set by the system for data communication. This inequality indicates that the probability of deviation from the baseline is less than or equal to $\frac{\lambda_{j}+H}{4}$. In the practical Internet of Vehicles (IoV), as shown in Figure \ref{figures:pvN}, the process of implementing the DSQL algorithm involves several steps: First, deploy Q-learning-based data communication and differential privacy mechanisms on the vehicle side, and a multi-layer perceptron on the edge server side. Next, optimize energy consumption to enhance the efficiency of Q-learning-based data communication, and utilize the multi-layer perceptron on the edge server to improve data processing capabilities. Finally, apply the differential privacy mechanism on the vehicle side to protect the vehicle's reward value and privacy information, thus achieving efficient and reliable communication between vehicles.

First, the vehicle transmits data by Q-learning; second, the data is processed by the multi-layer perceptron, so that the transmission efficiency of data is optimized; finally, the data is scheduled in higher efficiency for the vehicular network, and the secure communication between vehicles is guaranteed. For the Q-learning-based data scheduling, suppose $NM$ is the total vehicles, $M_{i,j}$ is the power consumption, and then the algorithm's time complexity is $\mathrm{O}\left(NM \cdot M_{i,j} \right)$. 

    \section{Experimental Setup}\label{section:experimental setup}

To evaluate the performance of DSQL, the state-of-the-art Wu's scheme \cite{wu2020collaborative}, adaptive priority scheduling (AdPS) \cite{sharma2020adps}, Shinde's scheme \cite{ shinde2022collaborative }, TDPP\cite{ zhu2023time } and multinomial recurrent neural network (MRNN) \cite{ safavat2023improved } are selected as the benchmark. Six evaluation metrics are adopted, which are accuracy, connectivity degree, travel expenses, transmission delay, probability of privacy leakage, and probability of malicious node attacks.

\subsection{Datasets}	

We use Veins and SUMO to establish vehicle movement models on the road to simulate the communication scenarios between vehicles and vehicles and between vehicles and infrastructure. The data sets on KITTI~\footnote{http://www.cvlibs.net/datasets/kitti/index.php} and NuScenes~\footnote{https://www.nuscenes.org/} are used to perform simulation analysis in MATLAB. The experiment parameters are listed in Table \ref{PTS}. The simulation area is a rectangle with a length of $48km$ and a width of $36km$. Simulation time refers to an experiment conducted at 90 or 150-second intervals. We made a performance comparison among AdPS, Shinde's scheme, TDPP, MRNN, and Wu's solutions.

\begin{table}
	\caption{Simulation Parameter Settings}\label{PTS}\centering
	\scalebox{1.0}{
		\begin{tabular}{l|c}
			\hline
			\hline
			The Minimum Speed & $12 m/s$    \\
			\hline
			The Maximum Speed & $35 m/s$    \\
			\hline
			The Size of Packets & $2 k$ bytes    \\
			\hline
			Area of Simulation & $36.0 km$, $48.0 km$   \\
			\hline
			Simulation Time & $90 s$, $150 s$   \\
			\hline
			Range of Communication & $2 km$    \\
			\hline
			Range of Interference & $1 km$     \\
			\hline
			Time slot & $1 ms$  \\
			\hline
			\hline
	\end{tabular}}
\end{table}

\subsection{Benchmark Methods}
%\red{other benchmarks? and experiment on neural network and q learning}

$\bullet$ \textbf{Adaptive Priority Scheduling (AdPS):} For efficient information transmission, AdPS \cite{sharma2020adps} uses fuzzy logic to estimate the deadline of the request and calculate the priority of the request message. The algorithm provides real-time data services in a heterogeneous traffic environment and ensures urgently requested services to achieve fairness among users.

$\bullet$ \textbf{Wu's scheme:} Wu's scheme\cite{wu2020collaborative} first uses reinforcement learning algorithms to find routes through the collaboration between the cloud-edge-end. Secondly, the overhead and delay of data communication are optimized through proactive and preemptive methods. Finally, through collaborative learning, fast and efficient communication routes are obtained in the vehicle edge computing environment.

$\bullet$ \textbf{Shinde's scheme:} To minimize the delay and energy consumption of the Internet of Vehicles, Shinde and Tarchi\cite{ shinde2022collaborative } define a collaborative Q-learning method, which utilizes V2I communication to enable multiple vehicles to participate in the training process of a centralized Q-agent and learn about the environment and potential offloaded neighbors through V2V communication, making better network selection and offload decisions.

$\bullet$ \textbf{TDPP:} To guard against privacy attacks, Zhu \emph{et al.}\cite{ zhu2023time } propose a learning model that utilizes a multi-agent system and a customized differential privacy mechanism to achieve efficient navigation in real-time traffic environments.

$\bullet$ \textbf{MRNN:} Safavat and Rawat\cite{ safavat2023improved } proposed an improved multinomial recurrent neural network (MRNN) classifier. First, the registered vehicle is logged in and authenticated. Then, the MRNN classifier is used to predict the movement of the authenticated vehicle, and the Ellipse Curve Cryptography (2CK-ECC) based on the Caesar Combination key is studied for secure data transmission at the same time.

AdPS and Wu's solutions study the problem of data scheduling in different ways. AdPS uses fuzzy logic to estimate the request deadline and calculate its priority so that vehicles can obtain adaptive priority scheduling services. Wu's solution uses collaborative learning among the cloud-edge end to obtain an efficient route for communication.

Shinde and Tarchi \cite{ shinde2022collaborative } proposed a collaborative Q-learning method, through V2I communication with multiple vehicles for centralized Q-agent training, and through V2V communication to make better network selection and offloading decisions; Zhu \emph{et al.} \cite{ zhu2023time } proposed a learning model, through a multi-agent system and a differential privacy mechanism to realize efficient navigation; and Safavat and Rawat\cite{ safavat2023improved } use a recurrent neural network to predict the movement of vehicles, and through the elliptic curve cryptography to achieve network security data transmission.

To obtain intelligent and efficient data communication, DSQL combines Q-learning and multi-layer perceptron methods in the vehicular network. In the data communication phase, the vehicle can obtain a better connection service owing to the cooperation between vehicles and the edge server. With the help of a multi-layer perceptron, data processing is made more efficient by narrowing the scope of data processing.

\subsection{Evaluation Metrics}	

%According to several studies related to mobile ad hoc networks (MANETs), the network speed and the bandwidth are considered as metrics for some work. However, the indicators are effective when infrastructure exists or the network's mobility is low. Because of the high mobility and the short connectivity of the vehicles, metrics of the network speed and the bandwidth is unrealistic. 
%The following metrics are considered to quantitatively analyze the performance of these three schemes.
The following six indicators are used to analyze the performance of the above schemes and the scheme proposed in this article.

$\bullet$ \textbf{Accuracy:} Accuracy represents the ratio of the number of correct predictions to the total number of predictions.

$\bullet$ \textbf{Travel Expenses (TE):} The cost of the vehicle from the place of departure to the destination.

$\bullet$ \textbf{Connectivity Degree:} The number of vehicles communicating with each other over some time.
%The number of vehicles connected within a certain communication range.

$\bullet$ \textbf{Transmission Delay (TD):} When data is received, the difference between the actual receiving time and the expected receiving time.

$\bullet$ \textbf{Probability of Privacy Leakage:} In the process of a malicious vehicle launching an attack, the number of times successfully obtaining the vehicle's privacy accounts for the proportion of the total number of attacks launched by the vehicle.

$\bullet$ \textbf{Probability of Malicious Node Attacks:} In the process of several malicious vehicles launching attacks, the number of times successfully hazard to vehicles accounts for the proportion of the total number of attacks launched by the vehicles.
%The difference between the expected time to get to the destination node and the actual time of arrival.

%$\bullet$ \textbf{Average Distance Deviation (ADD):} The average distance deviation between two vehicles in unit time.
%\begin{figure}
%	\centering	
%	{\includegraphics[width=2.6in,angle=0]{figures//Acc1.pdf}}
%	\caption{Accuracy under different distances.}\label{figures:acc}
%\end{figure}
\begin{figure}[t]
	\centering
	\subfigure[\label{figures:acc}
	KITTI.] 
	{\includegraphics[width=2.6in,angle=0]{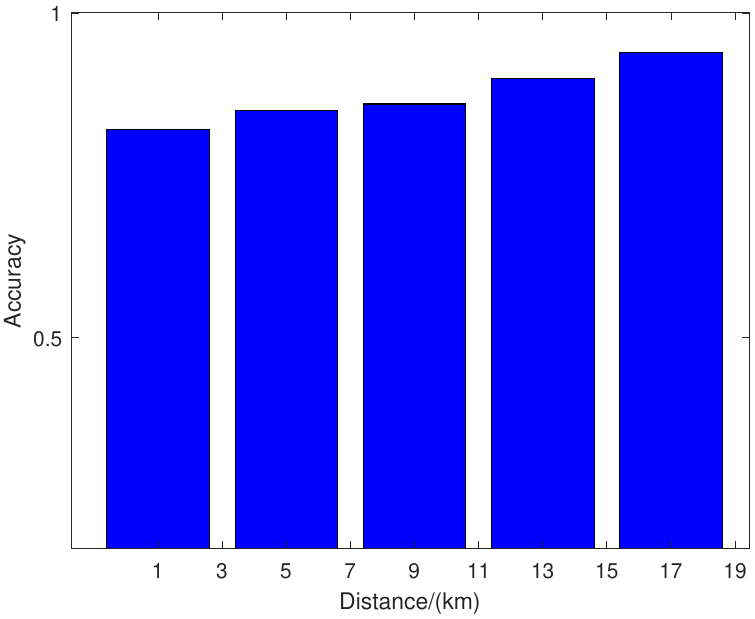}}
	\subfigure[\label{figures:acca}
	NuScenes.] 
	{\includegraphics[width=2.6in,angle=0]{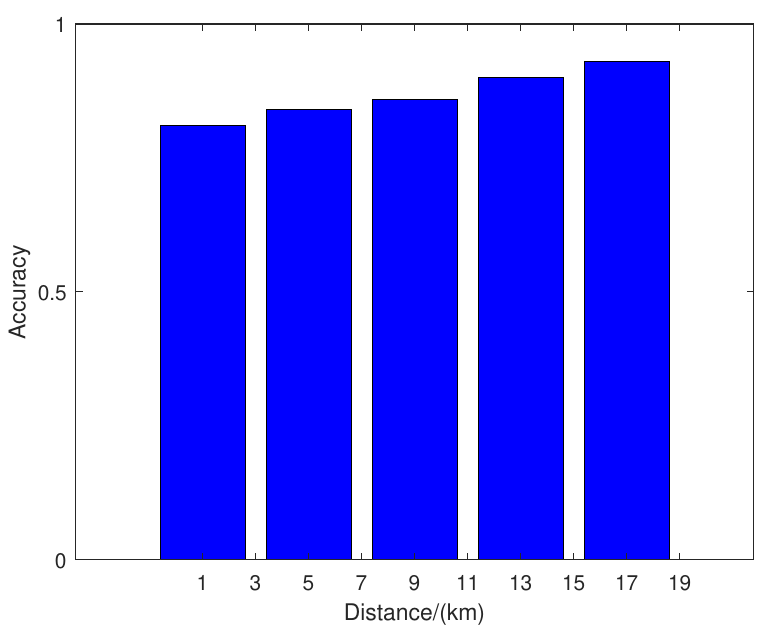}}
	\caption{Accuracy under different distances.}
\end{figure}
\begin{figure}[t]
	\centering
	\subfigure[\label{figures:add}
	KITTI.] 
	{\includegraphics[width=2.6in,angle=0]{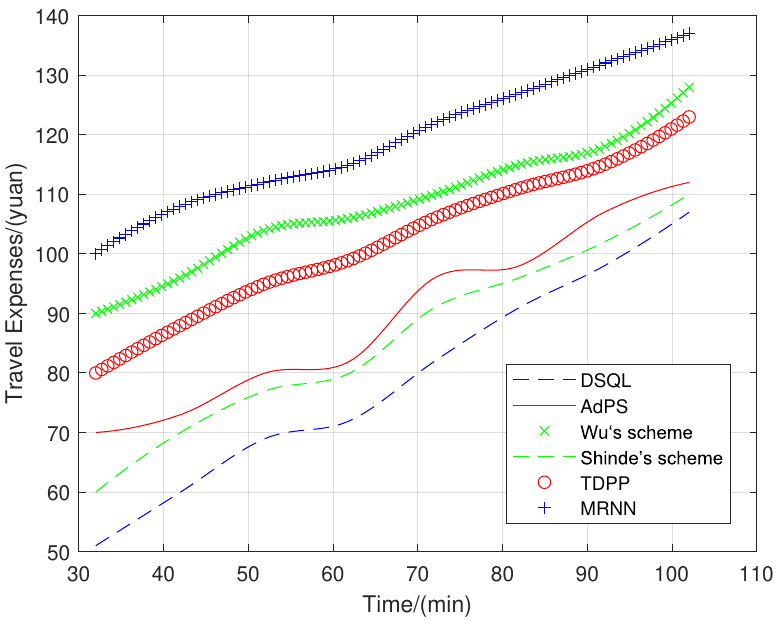}}
	\subfigure[\label{figures:traa}
	NuScenes.] 
	{\includegraphics[width=2.6in,angle=0]{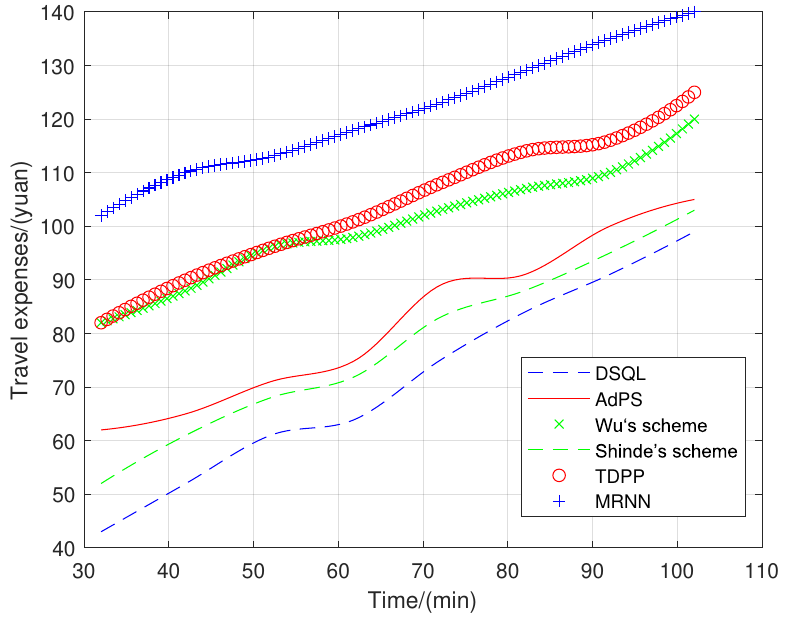}}
	\caption{Travel expenses under different time.}
\end{figure}
%\begin{figure}
%	\centering	
%	{\includegraphics[width=2.6in,angle=0]{figures//Expenses.pdf}}
%	\caption{Travel expenses under different time.}\label{figures:add}
%\end{figure}

\section{Performance Analysis}\label{section:results analysis}
%\red{the result graphs are too simple, avoid using these}
We analyze the performance of DSQL through six different indicators as follows.
% The experimental results show that the travel expenses of DSQL are better than that of Shinde's scheme, AdPS, TDPP, Wu's scheme, and MRNN, which shows that DSQL has better data scheduling performance. Moreover, the performance of connectivity and transmission delay of the proposed algorithm is better. 
\begin{figure}[t]
	\centering
	\subfigure[\label{figures:tdt}
	KITTI.] 
	{\includegraphics[width=2.6in,angle=0]{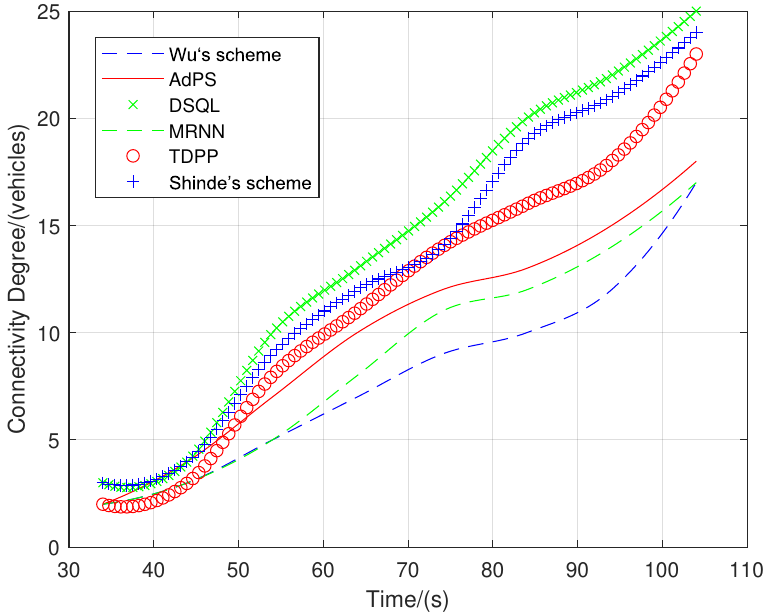}}
	\subfigure[\label{figures:conna}
	NuScenes.] 
	{\includegraphics[width=2.6in,angle=0]{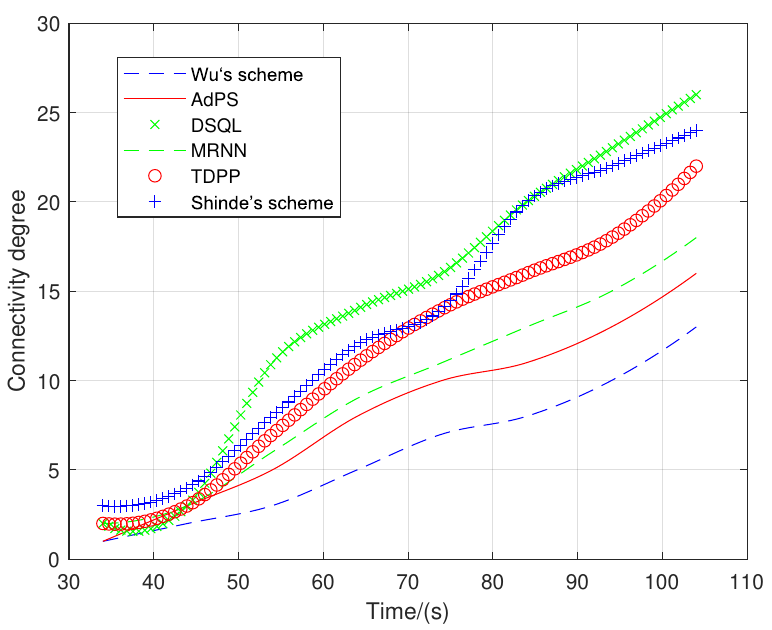}}
	\caption{Connectivity degree under different time.}
\end{figure}
%\begin{figure}
%	\centering	
%	{\includegraphics[width=2.6in,angle=0]{figures//conn.pdf}}
%	\caption{Connectivity degree under different time.}\label{figures:tdt}
%\end{figure}
\subsection{Accuracy}

The results of Q-learning are to make data transmission between vehicles more efficient and accurate. As mentioned in Section V, the data set used for training and testing comes from the widely used KITTI data set, which includes $6000$ types of data information from $3$ categories, $5000$ training images, and $1000$ test images. A different number of image categories are randomly assigned within each distance range. For example, $2000$ meters contain one category of images, while $4000$ meters contain $2$ categories, and the test set includes all three categories of image data. To observe the accuracy of the prediction, $300$ images are randomly selected from the test set in each test iteration within each range distance and the process is run 15 times, and the result of the accuracy is finally obtained, as shown in Figure \ref{figures:acc}. The accuracy of our algorithm has reached $0.82$, $0.85$, $0.87$, $0.89$, and $0.91$, and the result of data communication in the actual environment is acceptable.

%For this purpose we collected 1000 driving scenes in Boston and Singapore, two cities that are known for their dense traffic and highly challenging driving situations. The scenes of 20 second length are manually selected to show a diverse and interesting set of driving maneuvers, traffic situations and unexpected behaviors. The rich complexity of nuScenes will encourage development of methods that enable safe driving in urban areas with dozens of objects per scene. Gathering data on different continents further allows us to study the generalization of computer vision algorithms across different locations, weather conditions, vehicle types, vegetation, road markings and left versus right hand traffic.

The data set used for training and testing comes from the widely used NuScenes data set, which includes $1000$ driving scenes in Boston and Singapore, $800$ training images, and $200$ test images. The scenes of 20-second length are manually selected to show a diverse and interesting set of driving maneuvers, traffic situations, and unexpected behaviors. The rich complexity of NuScenes will encourage the development of methods that enable safe driving in urban areas with dozens of objects per scene, as shown in Figure \ref{figures:acca}. The accuracy of our algorithm has reached $0.86$, $0.88$, $0.91$, $0.94$, and $0.96$, and the result of data communication in the actual environment is acceptable.

% and \ref{figures:acca}
\subsection{Travel Expenses}

%Fig. \ref{figures:add} shows that our scheme achieves the best travel expenses among the above algorithms. AdPS has a lower travel expense, and Wu's scheme is the lowest. It is mainly because our scheme transmits the data by Q-learning, and the data is processed faster. AdPS scheme leverages the priority to schedule data, and the travel expense is lower than our scheme. Wu's scheme selects the vehicle through fuzzy logic, and the travel expense is higher than AdPS. However, our scheme not only considers Q-learning but also makes use of the neural network to process data faster. The travel expense of AdPS is lower than that of our scheme.
Figure \ref{figures:add} and Figure \ref{figures:traa} show that our scheme achieves the best travel expenses among the above algorithms, and MRNN’s travel expense is the worst. Shinde’s scheme is higher than AdPS, AdPS is higher than TDPP, and TDPP is higher than Wu's scheme. It is mainly because our scheme transmits the data by Q-learning, and the data is processed faster. Shinde’s scheme makes better network selection and offloading decisions through collaborative Q-learning, and the travel expense is higher than DSQL. AdPS scheme leverages the priority to schedule data, and the travel expense is higher than Shinde’s scheme. TDPP realizes efficient navigation by learning model, and the travel expense is higher than AdPS. Wu's scheme selects the vehicle through fuzzy logic, and the travel expense is higher than TDPP. MRNN achieves network security data transmission by recurrent neural networks and elliptic curve cryptography, and the travel expense is higher than Wu’s scheme. However, our scheme considers Q-learning, and differential privacy, and makes use of the multi-layer perceptron to process data faster. The travel expense of Shinde’s scheme is higher than that of our scheme.
\begin{figure}
	\centering
	\subfigure[\label{figures:nplrt}
	Transmission delay under different time.] 
	{\includegraphics[width=2.6in,angle=0]{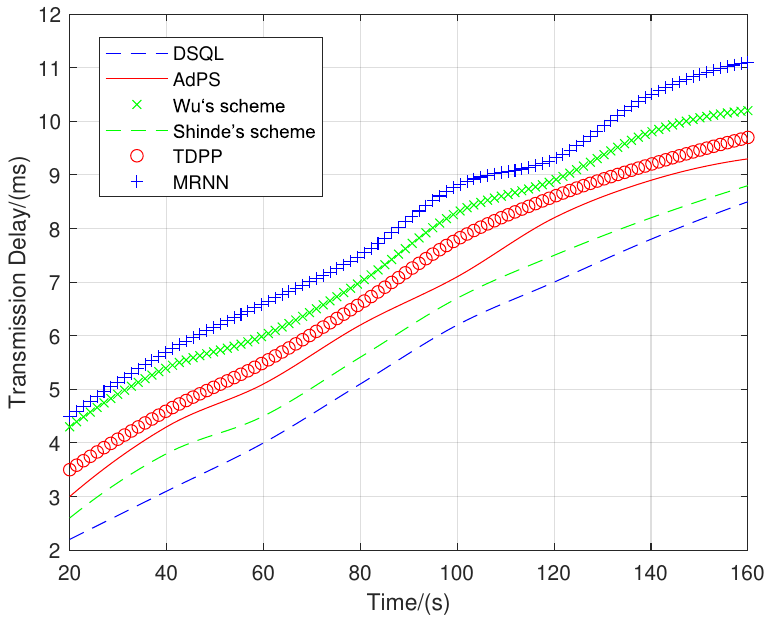}}
	\subfigure[\label{figures:nplrd}
	Transmission delay under different distances.] 
	{\includegraphics[width=2.6in,angle=0]{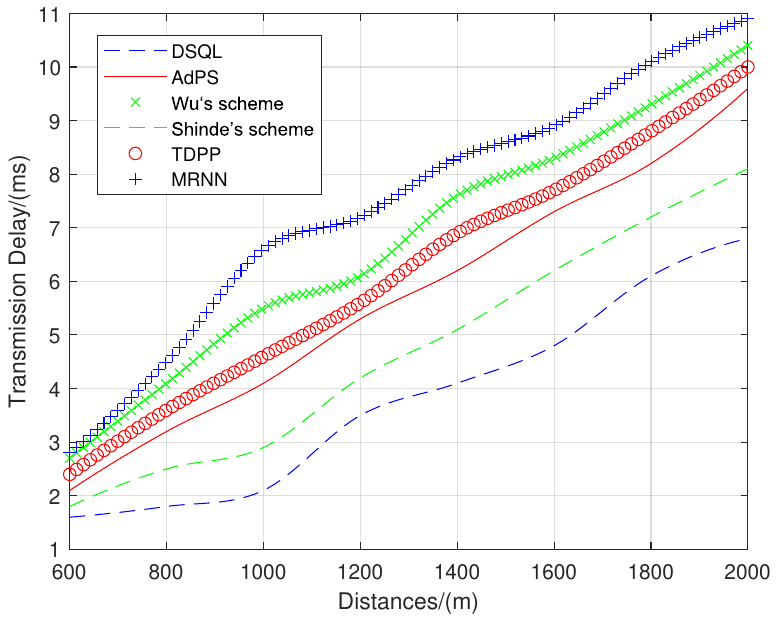}}
	\caption{Transmission delay under different time and distances (KITTI).}
\end{figure}
\begin{figure}
   \centering
   \subfigure[\label{xxx}
	Transmission delay under different time.] 
	{\includegraphics[width=2.6in,angle=0]{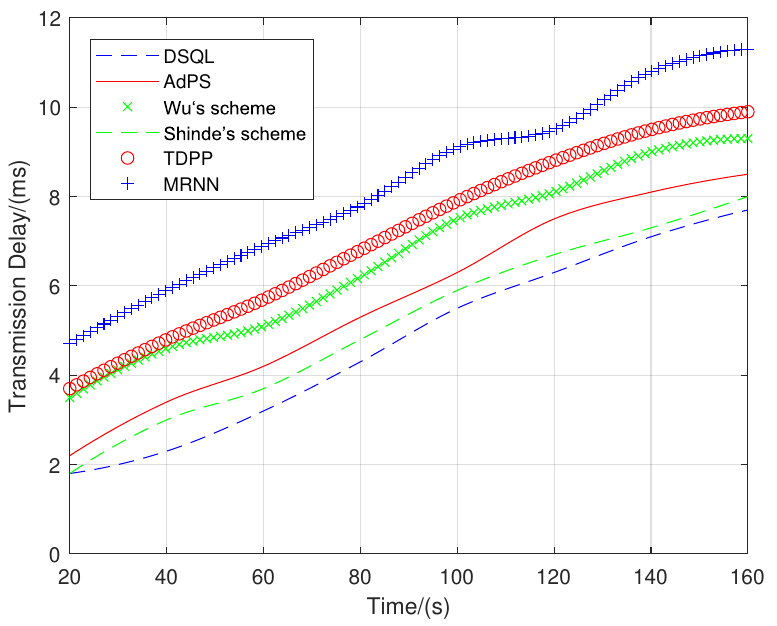}}
	\subfigure[\label{yyy}
	Transmission delay under different distances.] 
	{\includegraphics[width=2.6in,angle=0]{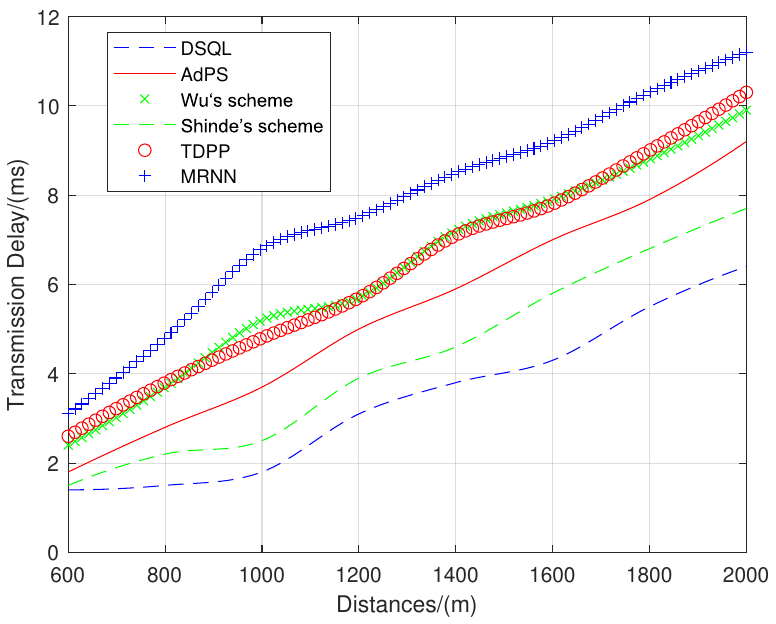}}
 \caption{Transmission delay under different time and distances (NuScenes).}
\end{figure}
\subsection{Connectivity Degree}

%Fig. \ref{figures:tdt} shows that our scheme achieves the highest connectivity degree compared with the benchmark algorithms. AdPS scheme has a lower connectivity degree. Wu's scheme's connectivity degree is the lowest. It is mainly because our scheme transmits the data by Q-learning. AdPS schedules data by adaptive priority and fuzzy logic is used for data services. So, its connectivity degree is lower than our scheme. While Wu's scheme uses Q-learning for traffic-intensive applications, it also considers packet forwarding, which slows down the information processing substantially. So, its connectivity degree is the lowest. 
Figure \ref{figures:tdt} and Figure \ref{figures:conna} show that our scheme achieves the highest connectivity degree compared with the benchmark algorithms. Shinde’s scheme is lower than DSQL, and TDPP is lower than Shinde’s scheme. AdPS is lower than TDPP, and MRNN is lower than AdPS. Wu's scheme's connectivity degree is the lowest. This is mainly because our scheme transmits the data by Q-learning. Shinde’s scheme makes offloading decisions and network selection through collaborative Q-learning, and the connectivity degree is lower than our scheme. TDPP uses a learning model to realize navigation and privacy protection of vehicles, and the connectivity degree is lower than Shinde’s scheme. AdPS schedules data by adaptive priority, and fuzzy logic is used for data services. The connectivity degree is lower than TDPP. MRNN leverages recurrent neural networks and elliptic curve cryptography to transmit data securely, and the connectivity degree is lower than AdPS. While Wu's scheme uses Q-learning for traffic-intensive applications. It also considers packet forwarding, which slows down information processing substantially. So, the connectivity degree is the lowest.
\subsection{Transmission Delay}

%Fig. \ref{figures:nplrt} and Fig. \ref{figures:nplrd} show that the transmission delay of DSQL is the lowest. As observed, the transmission delay increases as time increases. As the data transmitted by the vehicle and the resources consumed increase, the transmission delay of the three schemes will also keep rising. The main reason is as follows. The longer the transmission time is, the larger the transmission delay is. It can be seen from Fig. \ref{figures:nplrt} that DSQL has the best transmission delay performance, AdPS is second, and Wu’s scheme has the worst transmission delay performance. This is due to Q-learning between vehicles through enhanced neural networks to make communication more efficient. As observed in Fig. \ref{figures:nplrd}, the transmission delay goes up as the distance increases. The network will consume more resources, and the transmission delay of the above three schemes also keeps rising. The reason is that all three schemes have low transmission delays because of a small amount of data delivered at the initial stage. The transmission delay continues to grow due to the increase in distance and unstable factors caused by high-speed movement. These results demonstrate that the transmission delay of DSQL is better than the other two schemes. 
Figure \ref{figures:nplrt}, Figure \ref{figures:nplrd}, Figure \ref{xxx}, and Figure \ref{yyy} show that the transmission delay of DSQL is the lowest. As observed, the transmission delay increases as time increases. As the data transmitted by the vehicle and the resources consumed increase, the transmission delay of the six schemes will also keep rising. The main reason is as follows. The longer the transmission time is, the larger the transmission delay is. It can be seen from Figure \ref{figures:nplrt} and Figure \ref{xxx} that DSQL has the best transmission delay performance, Shinde’s scheme is second, AdPS is third, TDPP is fourth, Wu’s scheme is fifth, and MRNN has the worst transmission delay performance. This is due to Q-learning between vehicles through enhanced multi-layer perceptron to make communication more efficient. As observed in Figure \ref{figures:nplrd} and Figure \ref{yyy}, the transmission delay goes up as the distance increases. The network will consume more resources, and the transmission delay of the above six schemes also keeps rising. The reason is that all six schemes have low transmission delays because of a small amount of data delivered at the initial stage. The transmission delay continues to grow due to the increase in distance and unstable factors caused by high-speed movement. These results demonstrate that the transmission delay of DSQL is better than the other five schemes.
\begin{figure}
	\centering
	\subfigure[\label{figures:ppl}
	KITTI.] 
	{\includegraphics[width=2.6in,angle=0]{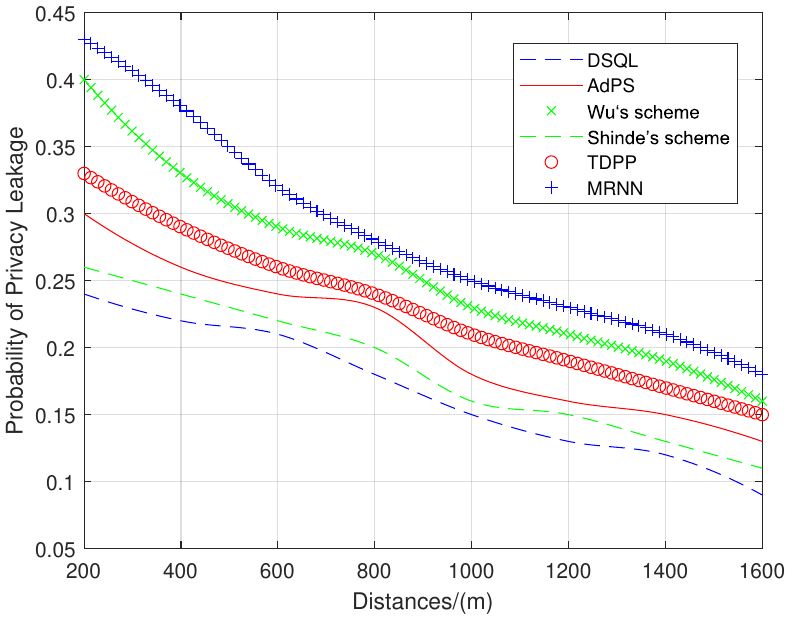}}
	\subfigure[\label{figures:prla}
	NuScenes.] 
	{\includegraphics[width=2.6in,angle=0]{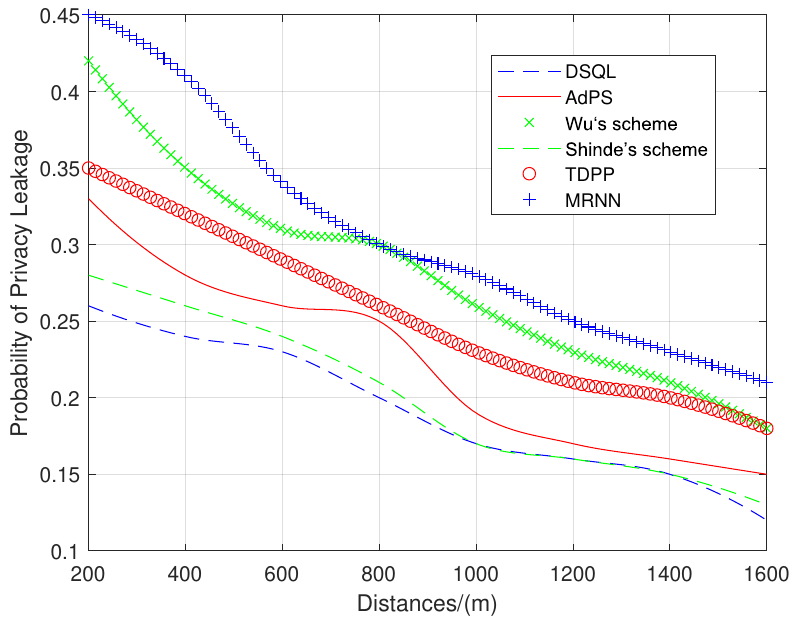}}
	\caption{Probability of privacy leakage under different distances.}
\end{figure}
%\begin{figure}
%	\centering	
%	{\includegraphics[width=2.6in,angle=0]{figures//ppl.pdf}}
%	\caption{Probability of privacy leakage under different distances.}\label{figures:ppl}
%\end{figure}

\subsection{Probability of Privacy Leakage}

%It can be seen from the figure \ref{figures:ppl} that the probability of privacy leakage of our scheme is the lowest, followed by Wu's scheme, and the probability of privacy leakage of AdPS is the highest. The probability of privacy leakage decreases with the increase of distance. As more and more communication resources are consumed in the network, the probability of privacy leakage of the above schemes decreases accordingly. It is because the closer the distance, the higher the probability of privacy leakage. The high-speed movement of the vehicle increases the distances, and brings uncertain environmental factors simultaneously, thereby reducing the probability of privacy leakage. This result shows that the probability of privacy leakage of our scheme is better than that of Wu's scheme and AdPS.

It can be seen from Figure \ref{figures:ppl} and Figure \ref{figures:prla} that the probability of privacy leakage of our scheme is the lowest, followed by Shinde's scheme, and AdPS is higher than Shinde’s scheme. TDPP is higher than AdPS, followed by Wu’s scheme, and the probability of privacy leakage of MRNN is the highest. The probability of privacy leakage decreases with the increase in distance. As more and more communication resources are consumed in the network, the probability of privacy leakage of the above schemes decreases accordingly. The closer the distance, the higher the probability of privacy leakage. The high-speed movement of the vehicle increases the distances, and brings uncertain environmental factors simultaneously, thereby reducing the probability of privacy leakage. This result shows that the probability of privacy leakage of DSQL is better than that of Shinde’s scheme, AdPS, TDPP, Wu's scheme, and MRNN.

\subsection{Probability of Malicious Node Attacks}

\begin{figure}
	\centering
	\subfigure[\label{figures:pmna}
	KITTI.] 
	{\includegraphics[width=2.6in,angle=0]{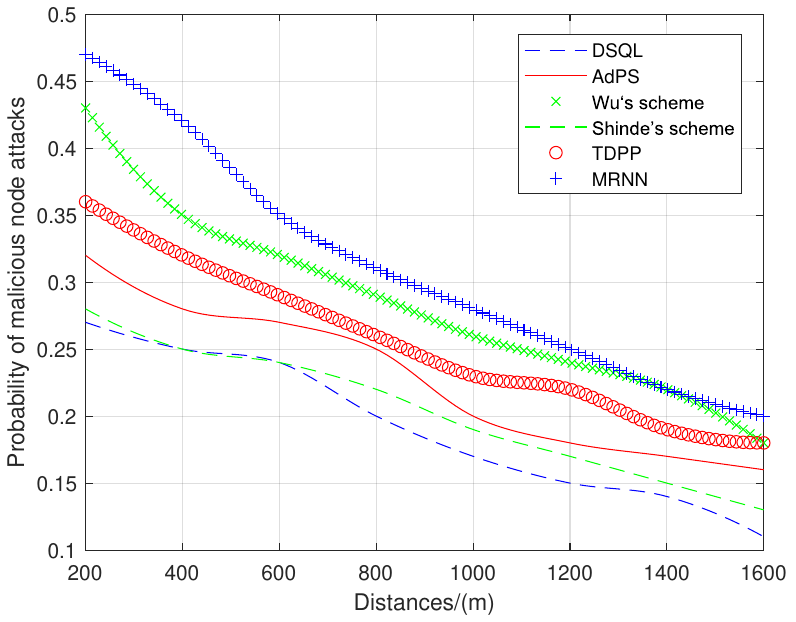}}
	\subfigure[\label{figures:mnaa}
	NuScenes.] 
	{\includegraphics[width=2.6in,angle=0]{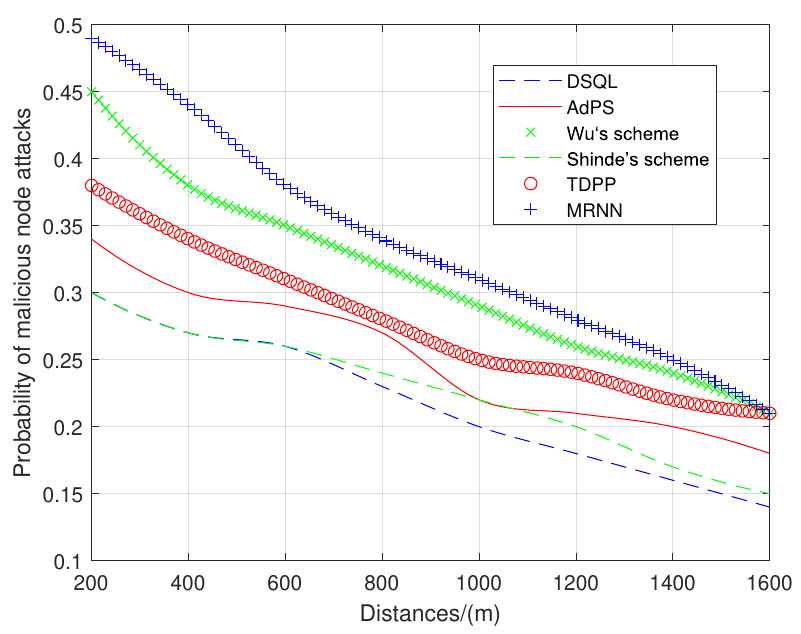}}
	\caption{Probability of malicious node attacks under different distances.}
\end{figure}

It can be seen from Figure \ref{figures:pmna} and Figure \ref{figures:mnaa} that the probability of malicious node attacks of our scheme is the lowest, followed by Shinde's scheme, and AdPS is higher than Shinde’s scheme. TDPP is higher than AdPS, followed by Wu’s scheme, and the probability of malicious node attacks of MRNN is the highest. The probability of malicious node attacks decreases with the increase in distance. This result shows that the probability of malicious node attacks on DSQL is better than that of Shinde’s scheme, AdPS, TDPP, Wu's scheme, and MRNN.

We can get some practical design inspiration from the simulation results. First, the accuracy of parameter settings and the rationality of assumptions, such as the range and difference of car speeds on urban roads and highways. Secondly, it is difficult for simulation experiments to be the same as the real environment. For example, there may be pedestrians or non-motor vehicles that suddenly cross the road in the real environment. Therefore, attention should be paid to the scope of the application and restrictions of the simulation results. Next, simulation is an iterative process. The model should be optimized according to the results and performance to make it more adaptable to the practical Internet of Vehicles environment. Finally, because it is difficult to cover all scenarios under extreme conditions, it is necessary to consider the scenarios under extreme conditions as comprehensively as possible, make more comprehensive predictions on possible damages, and prevent vehicles from experiencing different degrees of harmful incidents.

\section{Discussion of Machine Learning Based Data Scheduling for Vehicular Networks}\label{section:Discussion}

Given the extreme heterogeneity, dense deployment, dynamic characteristics, and strict quality of service requirements of network structures, machine learning has become the main solution for intelligent orchestration and management of networks. Through dynamic learning in uncertain environments, machine learning-supported channel estimation enables the full potential of ultra-wideband technology to be realized, and machine learning methods provide ultra-reliable, low-latency, and secure service guarantees for resource allocation and mobility management.
\subsection{Machine Learning Based Data Scheduling}

Generative AI technology can provide the ability for real data generation and advanced decision-making processes for vehicular networks, but it also faces challenges such as real-time data processing and privacy protection. To improve the quality of generative AI services, Zhang \emph{et al.} \cite{10506539} proposed a multimodal semantic perception framework that uses text and image data to create multimodal content, enhancing the usability and efficiency of vehicle systems. They also proposed a deep reinforcement learning-based resource allocation method to improve the reliability and efficiency of information transmission. As smart cities require capabilities in sensing, communication, computing, storage, and intelligence,  Chen \emph{et al.} \cite{10449899} proposed a paradigm of vehicles as a service, forming a network of mobile servers and communicators through vehicles to provide sensing, communication, computing, storage, and intelligent services for smart cities. They utilized potential use cases in smart cities and vehicular networks to construct a system architecture for the vehicle-as-a-service paradigm and pointed out future research directions for this paradigm. To enhance the generalization ability of the system, Tian \emph{et al.} \cite{10227873} proposed a unified framework based on the Transformer, VistaGPT, which includes a modular federation (MFoV) of vehicle Transformers for the automatic composition (AutoAuto) of autonomous driving systems. MFoV provides diversity and versatility to facilitate system integration, while AutoAuto leverages large language models to compose end-to-end autonomous driving systems. VistaGPT develops large language model-assisted transportation systems through capacity, scalability, and diversity, deploying scenario engineering systems.

The rapid development of generative AI technology has brought many opportunities for data scheduling in vehicular networks. By utilizing the multimodal capabilities of generative AI technology, data in vehicular networks can exhibit diversity, thereby providing a more reliable basis for decision-making in data transmission. By combining human feedback and multi-agent reinforcement learning, the accuracy and reliability of generative AI technology can be improved, making full use of vehicle-specific data, driving data, and environmental data to perceive traffic conditions in real-time, provide drivers with accurate and reliable driving information, and avoid congestion and unsafe incidents. In response to the issues of generative AI technology safety and vehicle driving safety, by studying attack methods and formulating defense mechanisms through V2X communication systems and privacy protection mechanisms, we can avoid the adverse information feedback caused by such attacks, which may lead to potential driving safety accidents. Therefore, generative AI technology can improve the accuracy and reliability of data scheduling in vehicular networks based on ensuring network security and driving safety, and provide high-quality service decisions for vehicle driving by utilizing vehicle data and traffic environment data.
\subsection{The Impact of Mobility Model}

To predict short-term traffic in the Internet of Vehicles, Chen \emph{et al.} \cite{10.1145/3430505} proposed a new data dissemination scheme. First, a three-layer network architecture was proposed to reduce communication overhead and deep learning was used to predict short-term traffic conditions. Then, a two-dimensional delay model was constructed to assign road section weights. Through global weight information, an ant colony optimization algorithm was used to find the path with the minimum delay. To utilize the idle computing resources of surrounding vehicles, Chen \emph{et al.} \cite{9684560} proposed a distributed multi-hop task offloading decision model, including a candidate vehicle selection mechanism and a task offloading decision algorithm. Because of the impact of different hop numbers on task completion, neighboring vehicles within the k-hop communication range are selected as candidate vehicles. The offloading problem is transformed into a constrained generalized allocation model and solved by a greedy algorithm and a discrete bat algorithm. To meet QoS/QoE requirements at a low cost, Li \emph{et al.}\cite{8434345} proposed a new Internet of Vehicles architecture to jointly optimize network, cache, and computing resources to alleviate network congestion. The programmable control principle of software-defined networks is used to promote system optimization and resource integration. A joint resource management solution is proposed to minimize system costs by modeling services, vehicle mobility, and system status. As vehicles become connected entities and data centers, Labriji \emph{et al.} \cite{9328479} demonstrate the value of vehicle mobility and the value of combining such estimates with online algorithms that ensure service continuity for vehicles. They use the Lyapunov method to solve the problem, obtain a low-complexity and distributed algorithm, and evaluate the performance of the algorithm in a scenario with thousands of vehicles and densely deployed 5G base stations.

The mobility model of vehicles has an important impact on the Internet of Vehicles. The mobility model of vehicles minimizes delays by predicting short-term traffic conditions and uses the idle computing resources of surrounding vehicles to complete tasks. The mobility model of vehicles minimizes system costs under the joint optimization of network, cache, and computing resources, and is used in conjunction with online algorithms to ensure service continuity through interconnected physical vehicles and is evaluated in dense scenarios. Therefore, the mobility model of vehicles has a great impact on the Internet of Vehicles in terms of delay, cost, resource management, and service quality.

\section{Conclusions} \label{section:conclusion}

This paper proposes Q-learning-based data scheduling (DSQL) for vehicular social networks. First, the DSQL uses Q-learning for data transmission. Secondly, the proposed algorithm processes the data quickly by the enhanced multi-layer perceptron to satisfy the requirement of the rapidity of the vehicles. The simulation results indicate that the DSQL is better than the state-of-the-art scheduling algorithm. 

In our future work, we will study the detection of abnormal driving behavior in intelligent transportation systems. The use of the blockchain method will enhance the privacy of vehicles, and this will also promote the development of data security in intelligent connected vehicles.

%\section*{Acknowledgements}

%This work was supported by the National Natural Science Foundation of China (Nos. 61772377, 61672257, 91746206) and the Science and Technology planning project of ShenZhen (JCYJ20170818112550194).

\bibliographystyle{IEEEtran} 
	\bibliography{Bibfile} 
	
\appendix 

In the DSQL algorithm proposed in this paper, the connectivity is defined as $\operatorname{Com}(x)=\frac{B_x \cdot V_r \cdot r_r}{V_s \cdot r_s}$ , where $B_x$ is the bandwidth of node $x$, $V_r$ is the speed of the vehicle receiving the message, $r_r$ is the data transmission rate of the receiving vehicle, $V_s$ is the speed of the sending vehicle, and $r_s$ is the transmission rate of the sending vehicle. The ratio of the speeds of the sending and receiving vehicles and the ratio of data rates are used for more accurate calculations. This is because the speed of the vehicles has a significant impact on their connectivity, a factor that was not considered in previous work. For example, the connectivity is determined by the ratio $C M(x)=\frac{\text { Num of hellos received from all } N B s}{\text { Num of hellos sent by all } N B s}=\frac{N_\tau}{N_s}$ \cite{wu2020collaborative} of the number of received messages to the number of sent messages, or by the ratio $C M(x)=\frac{h(x)}{\max _{y \in N_x} h(y)}=\frac{h(x)}{h^{\prime}(y)}$ of the antenna heights of the nodes. $\frac{\operatorname{Com}(x)}{C M(x)}=\frac{B_x \cdot V_r \cdot r_r}{V_s \cdot r_s} \cdot \frac{N_s}{N_r}>1 \text { if } \mathrm{B}_x \text { is large enough }$ or $\frac{\operatorname{Com}(x)}{C M(x)}=\frac{B_x \cdot V_r \cdot r_r}{V_s \cdot r_s} \cdot \frac{h^{\prime}(y)}{h(x)}>1 \text { if } \mathrm{B}_x \text { is large enough }$ , it can be concluded that increasing the bandwidth size can effectively improve the size of the connection degree through this ratio. Therefore, the connectivity of the algorithm proposed in this paper has better performance. 

$\operatorname{Re} d^{\prime}=\left\{\begin{array}{l}
\overline{\operatorname{Re} d}, \text { if } \mathrm{m} \text { is base station and } \mathrm{c} \text { is an edge node } \\
0, \text { otherwise }
\end{array}\right.$, 
where $\bar{Red} \in[0,1]$ is allocated by the BS according to the number of vehicles connected to the BS. 

$\operatorname{Re} d^{\prime}=\left\{\begin{array}{l}
1, \text { if } \mathrm{m} \text { is an edge node } \\
0, \text { otherwise }
\end{array}\right.$ .

In Q-learning-based data communication, we incorporate the amount of data processed into the reward design, where a higher reward value is assigned to stronger data processing capabilities, ultimately achieving higher data communication efficiency. As seen in the calculation of reward values during V2I and V2V communications

$\operatorname{Re} d=\left\{\begin{array}{l}
\overline{\operatorname{Re} d} \cdot D_n^{\text {local }}, \text { if } m \text { is base station } \\
0, \text { otherwise }
\end{array}\right.$ 
and

$\operatorname{Re} d=\left\{\begin{array}{l}
D_n^{\text {local }}, \text { if } m \text { is an edge node } \\
0, \text { otherwise }
\end{array}\right.$ . 

$\text { if } m \text { is base station and c is an edge node, } \frac{\operatorname{Re} d}{\operatorname{Re} d^{\prime}}=\frac{\overline{\operatorname{Re} d} \cdot D_n^{\text {local }}}{\operatorname{Re} d}=D_n^{\text {local }}$,

$\text { if } m \text { is an edge node, } \frac{\operatorname{Re} d}{\operatorname{Re} d^{\prime}}=\frac{D_n^{\text {local }}}{1}=D_n^{\text {local }}$. From this ratio, we can see that if the amount of processed data is large enough, our reward value will be larger.

\begin{equation}
\begin{aligned}\hat{R}=\min \left(1, \frac{P D_{t h}}{\frac{S_{p t h} * N_{u e}}{C B_{u l}}+\frac{S_{p h t} * N_{u e}}{C B_{c l}}+P D_{b s}}\right)\end{aligned},
\end{equation}
where $PD_{th}$ , $S_{pkt}$ , $N_{ue} $,$ CB_{ul}$ , $CB_{dl}$ , $PD_{bs}$ are the delay requirement, packet size, number of user devices, uplink bandwidth, down link bandwidth, and processing delay at the base station.
\begin{equation}
\begin{aligned}
\hat{R}=\min \left(1, \frac{P D_{t h}}{\frac{S_{p t t}}{C B_{11 p} \times H R R}+P D_{11 p}}\right)
\end{aligned},
\end{equation}
where $CB_{11p}$ is the bandwidth of IEEE 802.11p (27 Mbps) and HRR is the hello reception ratio between two neighbors. $PD_{11p}$ is the processing delay at each vehicle, which includes all the contention delay, the retransmission delay due to packet collisions, and the computing delay.

When using a cellular interface, factors such as the amount of data processed and latency are considered in calculating the reward value $\hat{\operatorname{Re}} d=\min \left(1, \frac{P D_{t h}+D_n^{\text {local }} \cdot T_{\text {persistent }}}{C B+P D_{b s}}\right)$; when using the IEEE 802.11P communication protocol, factors such as the amount of data processed, latency, and the probability of encountering obstacles are considered in calculating the reward value $\hat{R e} d=\min \left(1, \frac{P D_{\text {th }} \cdot T_{\text {persistent }}}{\frac{P S}{C B_{11 p} \cdot D_n^{\text {local }} \cdot H R R} \cdot P_{\text {obstacle }}+P D_{b s}}\right)$. Within a certain range of travel, the performance between vehicle latency and data processing quantity is balanced. 

\begin{equation}
\begin{aligned}
& \frac{\hat{R}}{\hat{\operatorname{Re} d}}=\frac{\frac{P D_{t h}}{\frac{S_{p k t} * N_{ue}}{C B_{ul}}+\frac{S_{p k t} * N_{ue}}{C B_{d l}}+P D_{b s}}}{\frac{P D_{th}+D_n^{\text {local }} \cdot T_{\text {persistent }}}{C B+P D_{b s}}}\\&=\frac{PD_{th}}{\frac{S_{pkt}*N_{ue}}{CB_{ul}}+\frac{S_{pkt}*N_{ue}}{CB_{dl}}+PD_{bs}}\cdot \frac{CB+P D_{b s}}{P D_{t h}+D_n^{\text {local }} \cdot T_{\text {persistent }}}\\&=\frac{C B+P D_{b s}}{\left(\frac{S_{p l t} * N_{u e}}{C B_{u l}}+\frac{S_{p l t} * N_{u e}}{C B_{d l}}+P D_{b s}\right) \cdot\left(1+\frac{D_n^{l o c a l} \cdot T_{p e r s i s \text { tent }}}{P D_{t h}}\right)} \\
&
\end{aligned}
\end{equation}

From this ratio, we can see that the first factor of the denominator is a number greater than $PD_{bs}$, and the second factor is a number greater than 1. Therefore, the denominator is a number greater than $PD_{bs}$, which can be written in the form of $n \cdot P D_{b s}, \quad n>1$. Therefore, when the upload and download bandwidth is smaller, our reward value is larger, and the processing delay of the base station will be smaller.

\begin{equation}
\begin{aligned}
&
\frac{\hat{R}}{\hat{Red}}=\frac{\frac{P D_{t h}}{\frac{S_{p k t}}{C B_{11 p} \times H R R}+P D_{11 p}}}{\frac{P D_{t h}\cdot T_{\text {persistent }}}{\frac{P S}{C B_{11 p} \cdot D_n^{local} \cdot H R R} \cdot P_{o b s tac e}+P D_{b s}}}\\&=\frac{\frac{P S}{C B_{11 p} \cdot D_n^{\text {local }} \cdot H R R} \cdot P_{\text {obstacle }}+P D_{b s}}{P D_{\text {th }} \cdot T_{\text {persistent }}}\\&=\frac{P D_{t h}}{\frac{S_{p k t}}{C B_{11 p} \times H R R}+P D_{11 p}} \cdot \\&\frac{\frac{P S}{C B_{11 p} \cdot D_n^{\text {local }} \cdot H R R} \cdot P_{\text {obstacle }}+P D_{b s}}{\left(\frac{S_{p k t}}{C B_{11 p} \times H R R}+P D_{11 p}\right) \cdot T_{\text {persistent }}} \\
& =\frac{P S \cdot P_{\text {obstacle }}+P D_{b s} \cdot D_n^{\text {local }}}{\left(S_{p \text { kt }}+P D_{11 p} \cdot H R R \cdot C B_{11 p}\right) \cdot T_{\text {persistent }} \cdot D_n^{\text {local }}} \\
& =\frac{\frac{P S}{D_n^{\text {local }}} \cdot P_{\text {obstacle }}+P D_{b s}}{\left(S_{p i t}+P D_{11 p} \cdot H R R \cdot C B_{11 p}\right) \cdot T_{\text {persisitent }}} \\
&
\end{aligned}
\end{equation}

From this ratio, we can see that because $P_{obstacle}$ is a number less than 1, assuming that under the IEEE 802.11P communication protocol, $\left(S_{p k t}+P D_{11 p} \cdot H R R \cdot C B_{11 p}\right)$ is a constant, then the more data is processed, the smaller the processing delay of the base station, and the larger our reward value can be, which is also in line with the practical situation of data transmission decision-making during vehicle movement.

The energy consumption value $E_n^{\text {local }}=p_n^{\text {local }} \Delta t=\frac{B\left(f_n^{\text {local }}\right)^2}{d} \Delta t$ and the amount of data processed $D_n^{\text {local }}=\frac{f_n^{\text {local }} \Delta t}{d \cdot a d}$ are calculated considering the impact of distance, and compared to the previous energy consumption \cite{luo2020collaborative} value $E_n^{\text {local }}=p_n^{\prime \text { local }} \Delta t=k_1\left(f_n^{\text {local }}\right)^3 \Delta t$, a power reduction is $\frac{E_n^{\text {local }}}{E_n^{\prime\text {local }}}=\frac{p_n^{\text {local }} \Delta t}{p_n^{\prime\text {local }} \Delta t}=\frac{B\left(f_n^{\text {local }}\right)^2}{d \cdot k_1\left(f_n^{\text {local }}\right)^3}=\frac{B}{d \cdot k_1}<1 \text { if distance is large enough, }$ it can be concluded that increasing the distance can effectively reduce the amount of energy consumption through this ratio. Therefore, the travel cost for vehicles is reduced by considering the impact of distance and using a more energy-efficient calculation method.

In data processing based on multi-layer perceptrons, we first apply a sine function mapping to the input data to reduce the range of data processing. Our activation function is $g(x)=\frac{1}{1+e^{-\sin (x)}}$, which, compared to the classic activation function $g(x)=\frac{1}{1+e^{-x}},$ incorporates an additional sine function mapping of the independent variable. As shown in Figure 5, the data range is almost halved. We introduce the sine function accordingly into the design of the cost function $J(\theta)=-\frac{1}{m} \sum_{k_1=1}^n \sum_{k_2=1}^{\infty} y_{k_2}^{k_1} \sin \left(h_\theta\left(x^{(1)}\right)\right)_{k_2}-\frac{1}{m} \sum_{k_1=1}^n \sum_{k_2=1}^{\infty}\left(1-y_{k_2}^{k_1}\right) \sin \left(1-h_\theta\left(x^{(1)}\right)\right)_{k_2}$ to minimize it. By reducing the data range, we enhance data processing capabilities, enabling faster data transmission decisions and ultimately reducing transmission latency. Within a smaller data range, the accuracy of data processing is higher. Therefore, by applying a sine function mapping to the input data and reducing the range of processed data, we can decrease data transmission latency and improve the accuracy of data transmission.

In data privacy protection, we utilize differential privacy mechanisms for data privacy protection and introduce pseudonym entropy for setting interference conditions. Differential privacy mechanisms are used to protect reward value information for vehicle traffic-sensitive applications and delay-sensitive unicast applications, i.e. $\text {Reward}= \begin{cases}\operatorname{Red} & \text { Traffic-intensive Applications } \\ \hat{Red} & \text { Delay-sensitive Unicast Applications }\end{cases}$. When setting interference conditions, to more effectively defend against attacks from malicious nodes, we introduce pseudonym entropy to limit the probability of deviation from real data within the range of pseudonym entropy \cite{ kang2017privacy}, i.e.,$\operatorname{Pr}\left[\hat{X}_j \neq x_j^*\right] \leq \frac{\lambda_j+H}{4}$, where $\lambda_j$ is a parameter set by the system during data communication. Experiments show that under the limitation of pseudonym entropy, the probability of privacy leakage and attacks from malicious nodes can be reduced by adjusting random parameters.

\end{document}